\documentclass[10pt,journal,compsoc]{IEEEtran}

\ifCLASSOPTIONcompsoc
  \usepackage[nocompress]{cite}
\else
  \usepackage{cite}
\fi
\usepackage[resetlabels]{multibib}
\newcites{appendix}{References}

\ifCLASSINFOpdf
  \usepackage[pdftex]{graphicx}
\else
  \usepackage[dvips]{graphicx}
\fi
\usepackage{amsmath}

\DeclareMathOperator*{\argmax}{argmax\,}
\usepackage{amssymb}
\usepackage{pifont}

\usepackage{algorithm}
\usepackage{algorithmicx}
\usepackage{algpseudocode}
\algrenewcommand\algorithmicrequire{\textbf{Input:}}
\algrenewcommand\algorithmicensure{\textbf{Output:}}
\usepackage{caption}
\usepackage{subcaption}
\usepackage{tablefootnote}
\usepackage{multirow}

\usepackage{stfloats}
\usepackage[hyphenbreaks]{breakurl}
\usepackage[hyphens]{url}

\usepackage{enumitem}
\setlist[itemize]{leftmargin=0.5cm,rightmargin=0.0cm}

\begin{document}
%
\title{Abstracting Concept-Changing Rules for\\Solving Raven's Progressive Matrix Problems}
%
%
%
%

\author{Fan Shi, Bin Li, and Xiangyang Xue 
\IEEEcompsocitemizethanks{\IEEEcompsocthanksitem The authors are with the Shanghai Key Laboratory of Intelligent Information Processing and the School of Computer Science, Fudan University,
Shanghai 200433, China\protect\\
E-mail: fshi22@m.fudan.edu.cn, \{libin, xyxue\}@fudan.edu.cn}
\thanks{Manuscript received April 19, 2005; revised August 26, 2015.}
\thanks{(Corresponding author: Bin Li.)}
}

%
%

\markboth{Journal of \LaTeX\ Class Files,~Vol.~14, No.~8, August~2015}%
{Shell \MakeLowercase{\textit{et al.}}: Bare Demo of IEEEtran.cls for Computer Society Journals}
%


\IEEEpubid{\begin{minipage}{\textwidth}\ \\[12pt]
	\copyright~2023 IEEE. Personal use of this material is permitted. Permission from IEEE must be obtained for all other uses, including reprinting/republishing this material for advertising or promotional purposes, collecting new collected works for resale or redistribution to servers or lists, or reuse of any copyrighted component of this work in other works. This work has been submitted to the IEEE for possible publication. Copyright may be transferred without notice, after which this version may no longer be accessible.
\end{minipage}}


\IEEEtitleabstractindextext{%
\begin{abstract}
  The abstract visual reasoning ability in human intelligence benefits discovering underlying rules in the novel environment. Raven's Progressive Matrix (RPM) is a classic test to realize such ability in machine intelligence by selecting from candidates. Recent studies suggest that solving RPM in an answer-generation way boosts a more in-depth understanding of rules. However, existing generative solvers cannot discover the global concept-changing rules without auxiliary supervision (e.g., rule annotations and distractors in candidate sets). To this end, we propose a deep latent variable model for \textbf{C}oncept-changing \textbf{R}ule \textbf{AB}straction (CRAB) by learning interpretable concepts and parsing concept-changing rules in the latent space. With the iterative learning process, CRAB can automatically abstract global rules shared on the dataset on each concept and form the learnable prior knowledge of global rules. CRAB outperforms the baselines trained without auxiliary supervision in the arbitrary-position answer generation task and achieves comparable and even higher accuracy than the compared models trained with auxiliary supervision. Finally, we conduct experiments to illustrate the interpretability of CRAB in concept learning, answer selection, and global rule abstraction.
\end{abstract}

}

\maketitle

\IEEEdisplaynontitleabstractindextext

%
\IEEEpeerreviewmaketitle

\IEEEraisesectionheading{\section{Introduction}\label{sec:introduction}}

\begin{figure}[!t]
  \centering
  \includegraphics[width=2.9in]{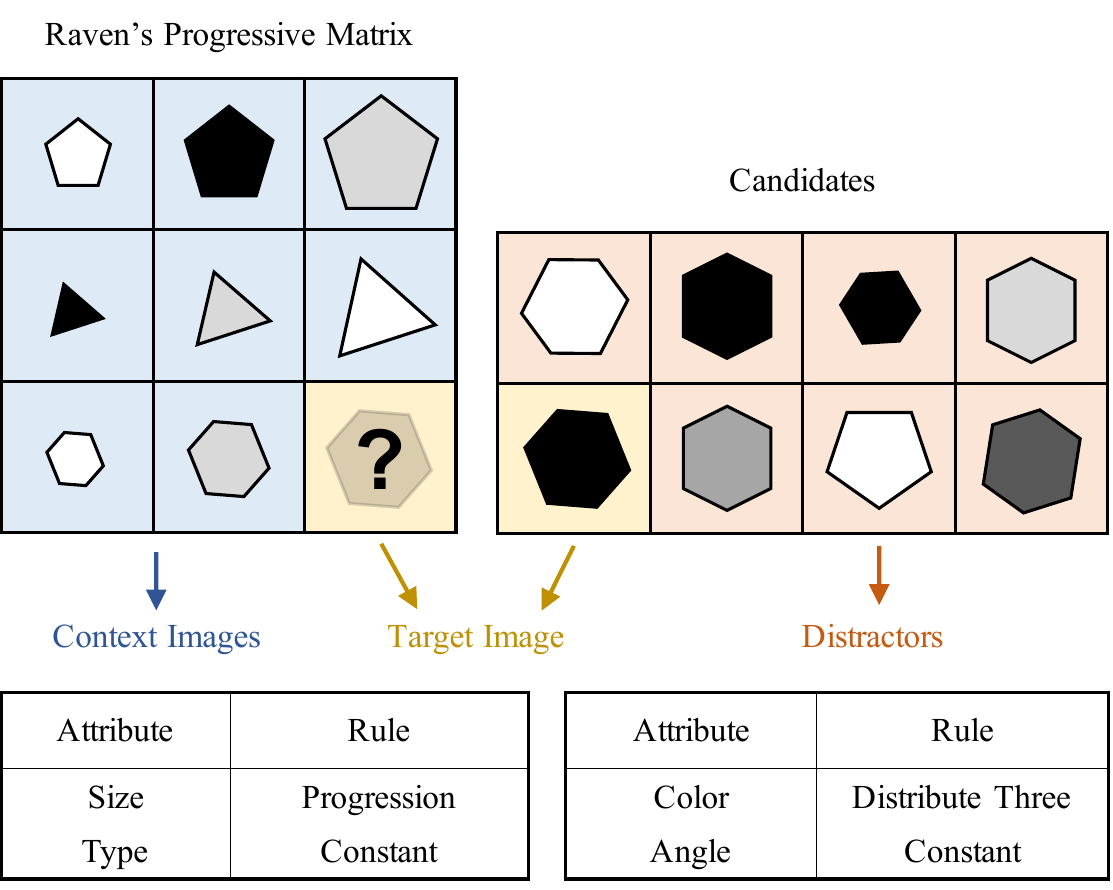}
  \caption{\textbf{An example of RPM}. The RPM consists of the context images (blue) and the target image (yellow) that is removed in the test. The candidate set (right) contains the removed target image and seven distractors (orange). We list the abstract rule on each attribute at the bottom.}
  \label{fig:rpm}
\end{figure}

\IEEEPARstart{T}{he} ability of abstract visual reasoning is critical for humans to understand underlying abstract rules and handle new problems \cite{cattell1963theory, zhuo2021effective, malkinski2022deep}. Raven's Progressive Matrix (RPM) \cite{raven1938raven} is a well-known abstract visual reasoning test used as an indicator of human intelligence \cite{snow1984topography, carpenter1990one, malkinski2022deep}. An RPM is a $3 \times 3$ matrix of images whose attributes (e.g., the size, shape, and color of objects) change with specific abstract rules. Figure \ref{fig:rpm} shows an RPM with monotonically increased object size. The bottom-right image is removed from the RPM to create the candidate set with seven distractors. In tests, participators should observe the context images, discover the underlying rules on the matrix, and choose the removed target image in the candidate set. The problem-solving process involves two sides of human cognition \cite{lovett2010structure}. A perception process converts input visual signals into abstract concepts (i.e., recognizing the attributes that change with abstract rules). A reasoning process parses the changing patterns of attributes and abstracts the global underlying rules shared among RPMs.

Abstract reasoning participates in various cognitive processes like number analogies, paper folding, geometric analogies, and letter series \cite{snow1989implications, ashton2000fluid, gray2004neurobiology, silvia2012making}. Therefore, endowing machine learning models with human-like reasoning ability is critical to developing higher-intelligence systems \cite{chollet2019measure}. In intelligence tests, RPM is less influenced by the language skill of participants and highly correlated with fluid intelligence \cite{snow1984topography, anastasi1997psychological, malkinski2022deep}. Recently, many machine learning models have emerged to develop the abstract visual reasoning ability by solving RPM tests \cite{barrett2018measuring, wu2020scattering, hu2021stratified}, which can enhance the generalization ability of models and help the application of skills in novel environment \cite{barrett2018measuring}.

Existing RPM solvers realize abstract visual reasoning in two ways. The selective solvers fill each candidate at the missing position, scoring the filled matrix and selecting the candidate image having the highest score \cite{barrett2018measuring, wang2019abstract, wu2020scattering}. However, one can directly imagine the missing images by understanding the underlying rules on RPMs \cite{hua2020modeling, pekar2020generating}. It is argued that solving RPMs as a generative task can reflect the in-depth understanding of rules and reduce the possibility of shortcut learning \cite{mitchell2021abstraction} (i.e., models achieve high selection accuracy by observing only candidate images). Therefore, training models with generative tasks can be a good choice to achieve human-like abstract visual reasoning \cite{mitchum2010solve, becker2016preventing}. To this end, a series of generative solvers have emerged to challenge the answer-generation problem. But most of them need auxiliary supervision (e.g., rule annotations or distractors in candidate sets) and can only predict the bottom-right images \cite{pekar2020generating, yu2021abstract, zhang2021learning, zhang2021abstract}. Therefore, there is still a large gap between the abstract visual reasoning ability of machine learning models and humans \cite{chollet2019measure}.

As an initial attempt at arbitrary-position answer generation, LGPP \cite{shi2021raven} and CLAP \cite{shi2022compositional} emphasize the role of interpretable concepts \cite{steenbrugge2018improving, wu2020scattering} and conduct reasoning with concept-specific conditional generation processes, which can be trained without auxiliary supervision. However, they are designed regarding RPMs with only continuous attributes and can hardly tackle more challenging datasets like RAVEN. Besides, one can automatically categorize rules on observations and apply them in novel RPMs. As the early categorical learning of infants is mainly driven by unsupervised perception \cite{behl1996basic, quinn2002category}, representing and discovering global rules without or with little auxiliary information is valuable in machine intelligence \cite{orhan2020self}. Another study also emphasizes the productivity and systematicity of the rule-based form of reasoning \cite{sloman1996empirical}. Although the rules are explicitly modeled as latent variables, LGPP and CLAP cannot further abstract global rules shared on all RPMs. 

\begin{figure*}[t]
  \vskip 0.2in
  \centering
  \includegraphics[width=0.95\textwidth]{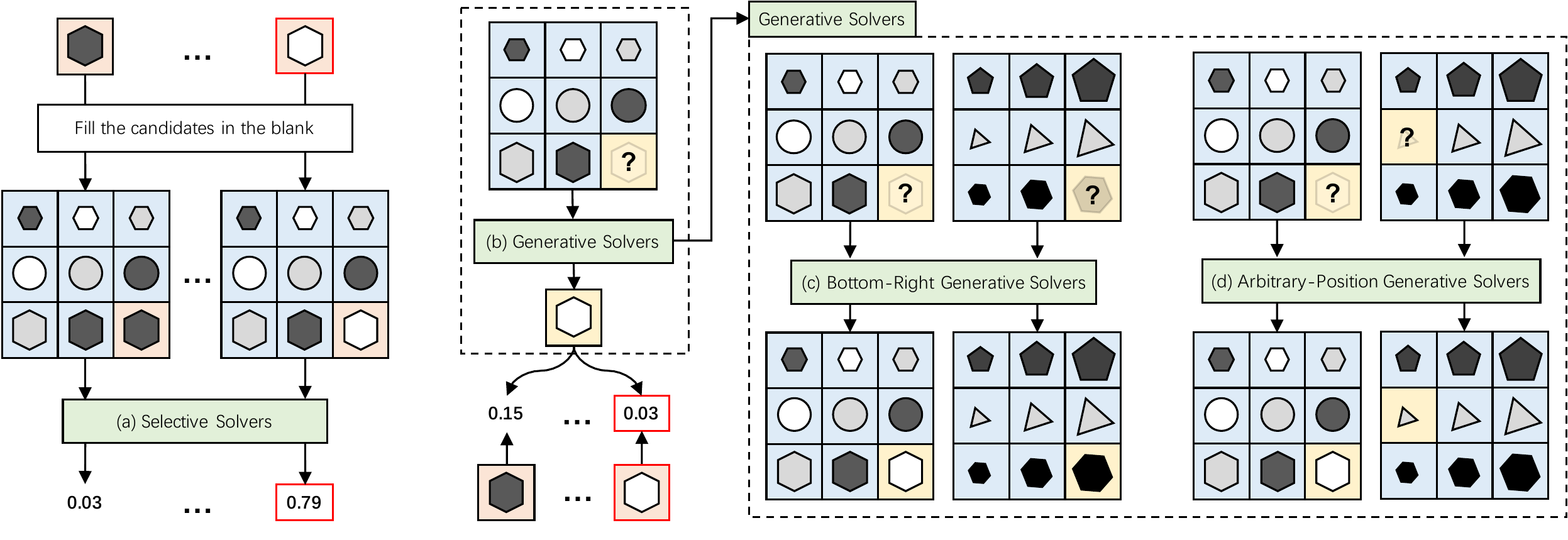}
  \caption{\textbf{The category of RPM solvers}. (a) \textit{Selective solvers} fill the candidates into the incomplete context panel and choose the candidate with the highest score as the answer. (b) \textit{Generative solvers} pick the candidate that is most similar to the prediction result. In generative solvers, (c) reasoning modules of \textit{bottom-right solvers} can only handle the missing image at the bottom-right position, and (d) \textit{arbitrary-position solvers} can generate the target at any position for matrix imputation.}
  \label{fig:related-works}
\end{figure*}

This paper proposes a deep latent variable model for Concept-changing Rule ABstraction (CRAB), which solves generative RPM tests without auxiliary supervision\footnote{Code is available at \url{https://github.com/FudanVI/generative-abstract-reasoning/tree/main/crab}.}. CRAB regards concepts and concept-specific rules as latent variables while employing an encoder and decoder to connect images and latent concepts. CRAB realizes abstract reasoning through concept-specific computational processes and acquires explicit representations of rules for target concept prediction. CRAB adopts the alternating knowledge update and knowledge-guided rule parsing processes to abstract the global rules shared on the dataset. The contributions of CRAB are summarized as follows.
\begin{itemize}
  \item \textbf{Arbitrary-position answer generation without auxiliary supervision.} CRAB is trained without auxiliary supervision to generate the missing image at arbitrary and multiple positions. In the experiments of bottom-right answer generation, CRAB achieves comparable or higher accuracy to the compared models trained with auxiliary supervision. We also require the models to generate answers at arbitrary and multiple positions, where CRAB outperforms the baselines trained without auxiliary supervision in most image configurations.
  \item \textbf{Concept learning and global rule abstraction.} CRAB represents image attributes with interpretable concepts in the latent space and parses rules for each latent concept. We propose an iterative learning process for CRAB to abstract global rules from the parsed concept-specific rules and update prior knowledge of rules. In the experiments, we visualize the latent concepts and explain how to choose or exclude a candidate. We also visualize the change of rule distributions in the training process and the representative RPMs sampled from different clusters of rules. The experimental results illustrate the interpretability of CRAB in concept learning, answer selection, and global rule abstraction.
\end{itemize}

\section{Related Work}

We categorize existing RPM solvers into \textit{selective solvers} and \textit{generative solvers} according to the way of producing answers in Figure \ref{fig:related-works}. Selective solvers choose answers by filling each candidate in the context panel and computing scores on the complete panel. And generative solvers choose answers by comparing candidates with the prediction results.

\subsection{Selective RPM Solvers}

In early studies, some selective solvers \cite{lovett2017modeling, little2012bayesian, lovett2010structure} fill each candidate to the given context to build eight candidate matrices, scoring them with human-designed representations, and select the candidate having the highest score. Recently, neural networks are widely taken as feature extractors due to their powerful non-linear approximation ability. Motivated by Relation Network \cite{santoro2017simple}, WReN \cite{barrett2018measuring} replaces human-designed representations with deep features and scores candidate matrices based on pairs of deep features.

To improve the abstract reasoning ability, existing solvers introduce disentangled representations \cite{steenbrugge2018improving, van2019disentangled}. VAE-WReN \cite{steenbrugge2018improving} replaces deep features in WReN with disentangled representations, and SCL \cite{wu2020scattering} introduces attribute-specific structure as additional inductive biases in score computation. Other models propose more specific architectures or training objectives as additional inductive biases. ARNe \cite{hahne2019attention} introduces multi-head attention and positional encodings to estimate scores \cite{vaswani2017attention}. CoPINet \cite{zhang2019learning} considers the exchange invariance of row permutation and proposes an NCE-like loss function \cite{gutmann2010noise} inspired by the contrast effect. LEN \cite{zheng2019abstract} extends WReN by encoding triplets of deep features and features of rows and columns, respectively. MLRN \cite{jahrens2020solving}, MRNet \cite{benny2021scale}, DCNet \cite{zhuo2021effective}, and SRAN \cite{hu2021stratified} utilize hierarchical structures on RPMs, including image features, row and column features, and abstract high-level features (e.g., the features obtained via arithmetical operations). 

The existing selective solvers have achieved outstanding selection accuracy in RPM tests. However, solving RPM problems in a fill-in-the-blank way can hardly explain the answer-generation ability in human abstract reasoning.

\subsection{Generative RPM Solvers}

Generative RPM solvers focus on the challenging task of answer generation. In recent years, deep generative models \cite{goodfellow2014generative, kingma2013auto} have achieved surprising performance in image generation. Niv et al. \cite{pekar2020generating} introduce a variational autoencoder (VAE) \cite{kingma2013auto} for feature extraction and image reconstruction, a context embedding network for feature prediction and a discriminator for adversarial training. ALANS \cite{zhang2021learning} and PrAE \cite{zhang2021abstract} extract symbolic features with Object CNN, realize abstract reasoning through algebraic abstraction or symbolic logic inference, and generate the missing image using the rendering engine. These \textit{bottom-right generative solvers} (Figure \ref{fig:related-works}c) can only generate answers at the bottom right of matrices and need auxiliary supervision in training. LoGe \cite{yu2021abstract} represents images through propositional variables, predicts representations of missing images by solving the maximum satisfiability problem (MAX-SAT), and reconstructs the answer from the predicted propositional variables. LoGe implements image encoding and decoding based on VQVAE \cite{van2017neural, razavi2019generating} and is trained without auxiliary supervision. But the reasoning module of LoGe can only generate bottom-right answers in simple configurations of I-RAVEN \cite{hu2021stratified}.

As \textit{arbitrary-position generative solvers}, LGPP \cite{shi2021raven} and CLAP \cite{shi2022compositional} capture concept-changing rules on RPMs using latent random functions to generate answers at arbitrary positions without auxiliary information in training. However, they can hardly adapt to RPMs with discrete attributes and rules and do not abstract global rules on attributes. In this paper, CRAB generates arbitrary-position answers by automatically learning interpretable concepts and global concept-specific rules, which is applicable to discrete RPMs such as RAVEN \cite{zhang2019raven} and I-RAVEN \cite{hu2021stratified}.

\section{Method}

CRAB solves the problem of arbitrary-position matrix imputation on RPMs. We will introduce the generative and inference processes that realize abstract reasoning, including the stage of concept learning, concept-specific rule parsing, and image generation. Finally, we provide details about module design, parameter learning, and global rule abstraction.

\subsection{Notations and Problem Definition}

We denote an RPM as $\boldsymbol{x} = \{\boldsymbol{x}_{n}\}_{n=1}^{9}$ in row-first order, i.e., $\boldsymbol{x}_{4}$ is the image at the position $(2,1)$ (the second row and first column). The whole matrix is split into mutually exclusive target images $\boldsymbol{x}_{T}$ and context images $\boldsymbol{x}_{C}$ through index sets $T$ and $C$. In this paper, the models are required to predict the target images $\boldsymbol{x}_{T}$ from the context $\boldsymbol{x}_{C}$. In addition to target prediction, CRAB attempts to discover global rules for each concept in the training process.

\subsection{Abstract Reasoning as Conditional Generation}

\begin{figure}[t]
  \centering
  \includegraphics[width=2.7in]{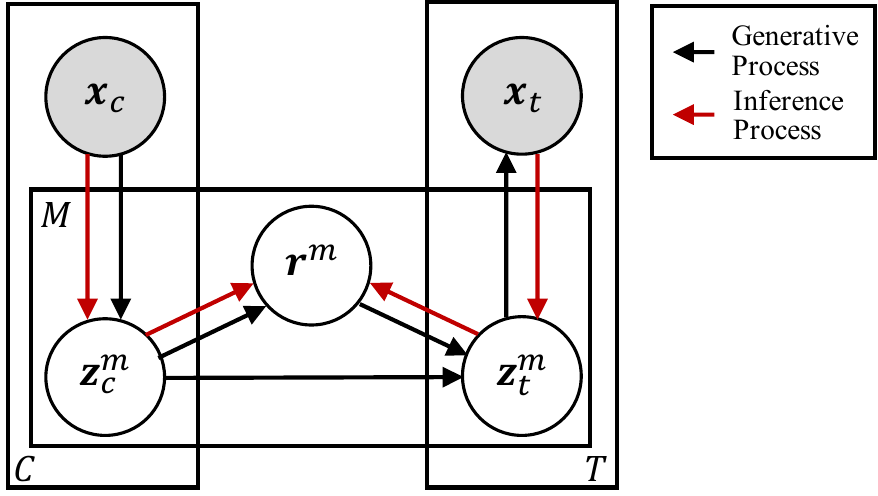}
  \caption{\textbf{The probabilistic graphical model of CRAB}. The circles with white (gray) backgrounds are latent variables (observations). The black-solid (red-dashed) lines indicate dependencies in the generative (inference) process.}
  \label{fig:gm}
\end{figure}

CRAB predicts target images by solving a conditional probability $p(\boldsymbol{x}_{T}|\boldsymbol{x}_{C})$, which is factorized into several simpler probabilities through interpretable latent variables $\boldsymbol{h}$ to reduce the complexity of reasoning. As a deep latent variable model, the encoder and decoder of CRAB connect high-dimensional images and low-dimensional latent variables, and the learnable mappings parse underlying rules in the latent space. The training objective of CRAB is to maximize the log-likelihood $\log p(\boldsymbol{x}_{T}|\boldsymbol{x}_{C})$, which has no closed-form solutions due to the complex learnable networks. To this end, we approximate the log-likelihood using the evidence lower bound (ELBO) $\mathcal{L}$ in variational Bayes \cite{kingma2013auto}:
\begin{equation}
  \begin{aligned}
    \mathcal{L} = \mathbb{E}_{q(\boldsymbol{h}|\boldsymbol{x})}\left[\log \frac{p\left(\boldsymbol{h}, \boldsymbol{x}_{T} | \boldsymbol{x}_{C}\right)}{q\left(\boldsymbol{h} | \boldsymbol{x}\right)}\right] \leq \log p\left(\boldsymbol{x}_{T} | \boldsymbol{x}_{C}\right).
  \end{aligned}
  \label{eq:svcl-elbo}
\end{equation}
The generative process $p(\boldsymbol{h},\boldsymbol{x}_{T}|\boldsymbol{x}_{C})$ deduces the latent variables and predicts target images from the given context. The inference process $q(\boldsymbol{h}|\boldsymbol{x})$ approximates the untractable posterior distribution $p(\boldsymbol{h}|\boldsymbol{x})$. The latent variables $\boldsymbol{h}$ consist of $\boldsymbol{z}=\{\boldsymbol{z}_{n}\}_{n=1}^{9}$ that encode attributes of each image (e.g., object size and shape) and $\boldsymbol{r}$ that represents rules on the entire matrix. This section will introduce the generative and inference processes to illustrate how CRAB conducts abstract reasoning based on interpretable latent variables.

\textbf{Concept Learning}. Compositionality benefits abstract reasoning in complex scenes \cite{steenbrugge2018improving, van2019disentangled, wu2020scattering}. Analyzing independent factors from visual signals is the key to understanding rules in the real world. For example, one parses the rules of time-variant position and time-invariant appearance to predict the object trajectory \cite{shi2022compositional}. CRAB introduces compositionality by decomposing $\boldsymbol{z}_{n}$ into $M$ independent concepts $\boldsymbol{z}_{n} = \{\boldsymbol{z}^{m}_{n}\}_{m=1}^{M}$. Then CRAB further decomposes the rule into concept-changing rules according to the concepts, i.e., $\boldsymbol{r}=\{\boldsymbol{r}^{m}\}_{m=1}^{M}$. The concept-specific rules bring two advantages: (1) by decomposing confounded image representations into concepts, we reduce the complexity of abstract reasoning; (2) the concept-specific rules can be categorized and shared on the dataset, which promotes the generalization ability of CRAB.

\textbf{Generative Process}. Figure \ref{fig:gm} illustrates the generative process that converts context images into latent variables and target images. CRAB factorizes $p(\boldsymbol{h},\boldsymbol{x}_{T}|\boldsymbol{x}_{C})$ as
\begin{equation}
  \begin{aligned}
    & p\left(\boldsymbol{h},\boldsymbol{x}_{T}|\boldsymbol{x}_{C}\right) = \prod_{t \in T} p\left(\boldsymbol{x}_{t}|\boldsymbol{z}_{t}\right) \\
    & \quad \quad \prod_{m=1}^{M} \left( p\left(\boldsymbol{z}^{m}_{T}|\boldsymbol{r}^{m},\boldsymbol{z}^{m}_{C}\right) p\left(\boldsymbol{r}^{m}|\boldsymbol{z}^{m}_{C}\right) \prod_{c \in C} p\left(\boldsymbol{z}^{m}_{c}|\boldsymbol{x}_{c}\right) \right) .
  \end{aligned}
  \label{eq:svcl-gen}
\end{equation}
The generative process includes the context encoding, abstract reasoning, and target decoding stages. $p(\boldsymbol{z}_{c}|\boldsymbol{x}_{c})$ is the context encoding stage that extracts concepts $\{\boldsymbol{z}_{c}^{m}\}_{m=1}^{M}$ for each context image. In the abstract reasoning stage, CRAB parses the concept-specific rules to predict concepts for target images. $p(\boldsymbol{r}^{m}|\boldsymbol{z}_{C}^{m})$ is the rule parsing process where learnable networks discover the global changing pattern of each concept on context images. We sample the rule representation $\boldsymbol{r}^{m} \sim p(\boldsymbol{r}^{m}|\boldsymbol{z}_{C}^{m})$ and generate the concepts of target images through $p(\boldsymbol{z}_{T}^{m}|\boldsymbol{r}^{m},\boldsymbol{z}_{C}^{m})$. At the end of the generative process, the target decoding stage $p(\boldsymbol{x}_{t}|\boldsymbol{z}_{t})$ takes the concepts of target images as inputs to reconstruct target images. The generative process involves the learnable encoder, decoder, rule parsers, and target predictors. Designing appropriate built-in structures in these learnable modules will promote the problem-solving ability of CRAB.

\textbf{Variational Inference}. CRAB leverages the entire matrix to parse concept-specific rules in the inference process. According to Figure \ref{fig:gm}, the variational distribution $q(\boldsymbol{h}|\boldsymbol{x})$ derives latent variables from both context and target images:
\begin{equation}
  \begin{aligned}
    q\left(\boldsymbol{h}|\boldsymbol{x}\right) = \prod_{m=1}^{M} \left( q\left(\boldsymbol{r}^{m}|\boldsymbol{z}^{m},\boldsymbol{x}\right) \prod_{n=1}^{9} q\left(\boldsymbol{z}^{m}_{n}|\boldsymbol{x}_{n}\right) \right) .
  \end{aligned}
  \label{eq:svcl-inf}
\end{equation}
$q(\boldsymbol{z}_{n}|\boldsymbol{x}_{n})$ is the first stage of the inference process, where the encoder converts context and target images into concepts. The concepts of different images are concatenated to parse the concept-specific rules through $q(\boldsymbol{r}^{m}|\boldsymbol{z}^{m})$. Since the concepts of target images are directly acquired from target images, the inference process does not involve the part of target prediction.

\begin{figure*}[t]
  \centering
  \begin{subfigure}[b]{0.6\textwidth}
    \centering
    \includegraphics[width=\textwidth]{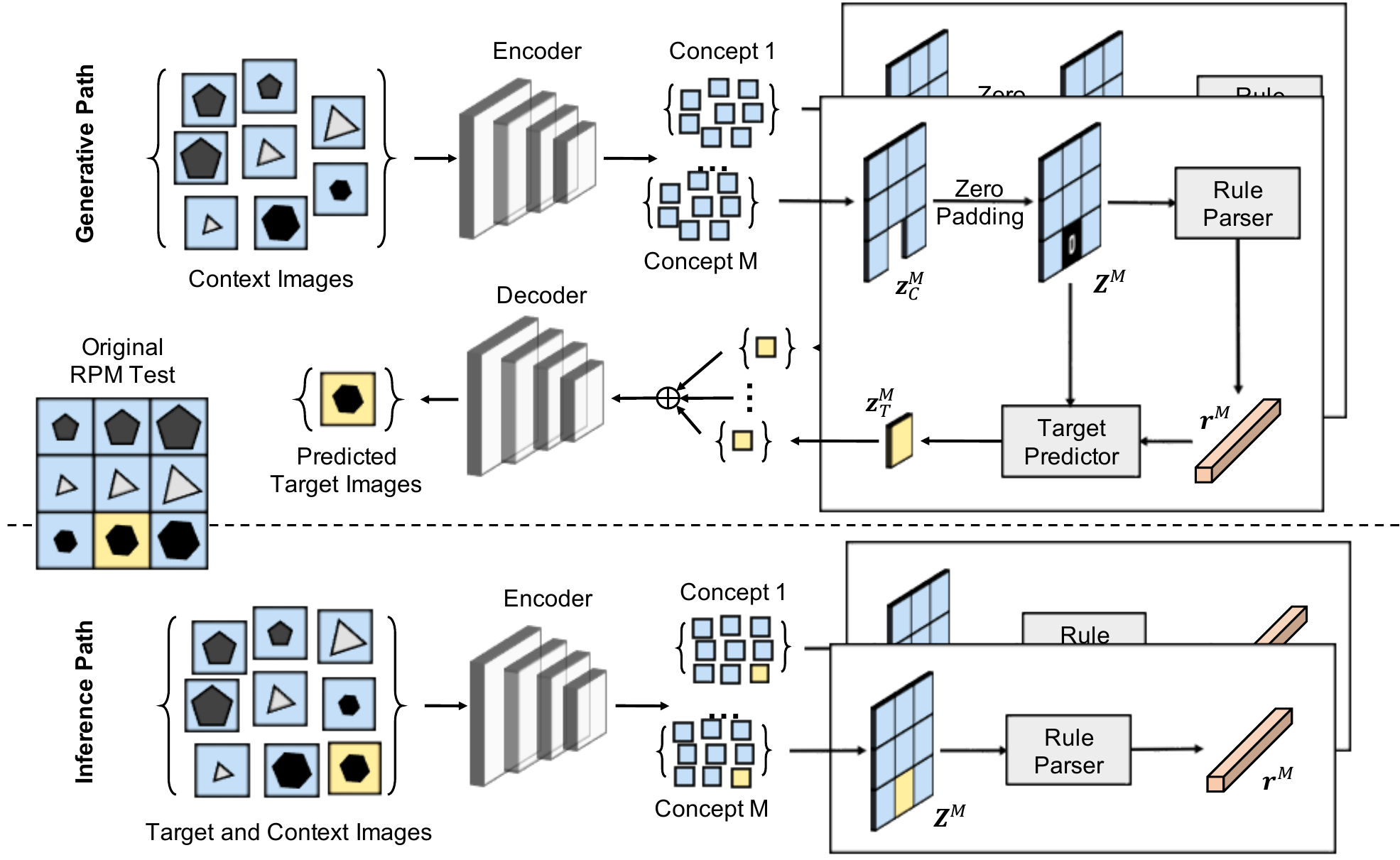}
    \caption{The generative and inference processes}
    \label{fig:svcl-process}
  \end{subfigure}
  \hfill
  \begin{subfigure}[b]{0.35\textwidth}
    \centering
    \vskip 0.1in
    \includegraphics[width=\textwidth]{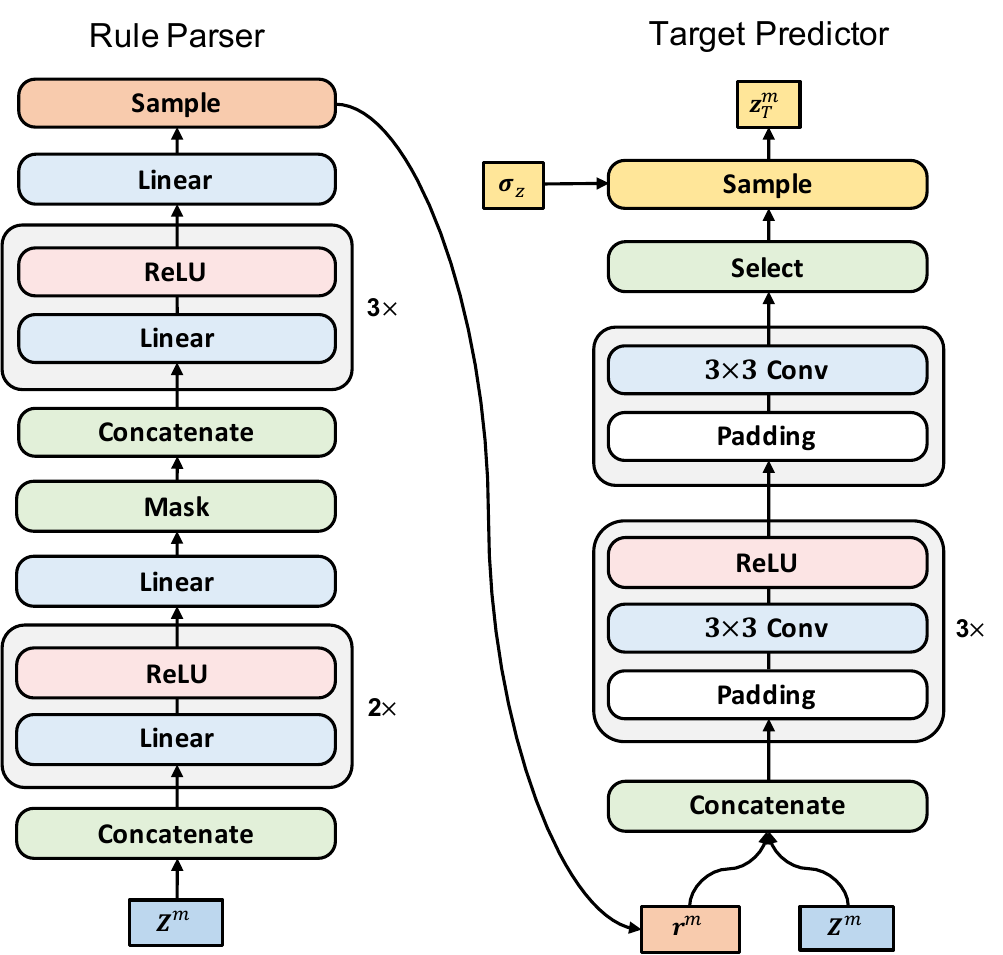}
    \vskip 0.2in
    \caption{The rule parser and target predictor}
    \label{fig:svcl-arch}
  \end{subfigure}
  \caption{\textbf{An illustration of the generative and inference processes}. (a) shows the computational path of the generative and inference processes. (b) are the architectures of concept-specific rule parsers (left) and target predictors (right).}
  \label{fig:svcl-process-arch}
\end{figure*}

\textbf{Parameter Learning}. Since neural networks are powerful and flexible to parameterize nonlinear transformations, CRAB models different probabilities in Equation \ref{eq:svcl-gen} and Equation \ref{eq:svcl-inf} with accordingly learnable neural networks. The encoder outputs the parameters of $q(\boldsymbol{z}_{n}|\boldsymbol{x}_{n})$ (e.g., the means and standard deviations in Gaussians). The decoder describes the inverse process $p(\boldsymbol{x}_{n}|\boldsymbol{z}_{n})$ that converts concepts into target images. CRAB provides concept-specific rule parsers and target predictors to capture the underlying rules in the abstract reasoning stage. The concept-specific rule parser and target predictor parameterize $q(\boldsymbol{r}^{m}|\boldsymbol{z}^{m})$ and $p(\boldsymbol{r}^{m}|\boldsymbol{z}_{C}^{m})$ and $p(\boldsymbol{z}_{T}^{m}|\boldsymbol{r}^{m},\boldsymbol{z}_{C}^{m})$ respectively. The rule parsers and target predictors can analyze the underlying rules from the concepts obtained from the complete or partial matrix.

The parameters of CRAB are optimized by maximizing the ELBO in Equation \ref{eq:svcl-elbo}, which is reformulated as
\begin{equation}
  \begin{aligned}
    \mathbb{E}_{q(\boldsymbol{h}|\boldsymbol{x})} \Big[ \log p\left(\boldsymbol{x}_{T}|\boldsymbol{h},\boldsymbol{x}_{C}\right) \Big] - \mathbb{E}_{q(\boldsymbol{h}|\boldsymbol{x})}\left[\log \frac{q\left(\boldsymbol{h}|\boldsymbol{x}\right)}{p\left(\boldsymbol{h}|\boldsymbol{x}_{C}\right)}\right].
  \end{aligned}
\end{equation}
According to Equations \ref{eq:svcl-gen} and \ref{eq:svcl-inf}, the first reconstruction term
\begin{equation}
  \begin{aligned}
    \mathcal{L}_{r} = \sum_{t \in T} \mathbb{E}_{q(\boldsymbol{h}|\boldsymbol{x})}\Big[\log p\left(\boldsymbol{x}_{t} | \boldsymbol{z}_{t}\right)\Big]
    \label{eq:svcl-elbo-lr}
  \end{aligned}
\end{equation}
measures the quality of reconstructed images by computing $\log p(\boldsymbol{x}_{t}|\boldsymbol{z}_{t})$ on each target image. The second regularizer $\mathcal{R}$ can be further factorized into $\mathcal{R}_{r}+\mathcal{R}_{t}+\mathcal{R}_{c}$ (see derivation in Appendix A) where
\begin{equation}
  \begin{aligned}
    \mathcal{R}_{r} &= \sum_{m=1}^{M} \mathbb{E}_{q(\boldsymbol{h}|\boldsymbol{x})} \left[ \log \frac{q\left(\boldsymbol{r}^{m}|\boldsymbol{z}^{m}\right)}{p\left(\boldsymbol{r}^{m}|\boldsymbol{z}_{C}^{m}\right)} \right], \\
    \mathcal{R}_{t} &= \sum_{m=1}^{M} \mathbb{E}_{q(\boldsymbol{h}|\boldsymbol{x})} \left[ \log \frac{q\left(\boldsymbol{z}_{T}^{m}|\boldsymbol{x}_{T}\right)}{p\left(\boldsymbol{z}_{T}^{m}|\boldsymbol{r}^{m},\boldsymbol{z}_{C}^{m}\right)}\right], \\
    \mathcal{R}_{c} &= \sum_{m=1}^{M} \sum_{c \in C} \mathbb{E}_{q(\boldsymbol{h}|\boldsymbol{x})} \left[ \log \frac{q\left(\boldsymbol{z}_{c}^{m}|\boldsymbol{x}_{c}\right)}{p\left(\boldsymbol{z}_{c}^{m}|\boldsymbol{x}_{c}\right)}\right].
  \end{aligned}
  \label{eq:svcl-elbo-regularizers}
\end{equation}
The \textbf{rule regularizer} $\mathcal{R}_{r}$ indicates the consistency of the parsed rules on concepts through the Kullback-Leibler (KL) divergence between $q(\boldsymbol{r}^{m}|\boldsymbol{z}^{m})$ and $p(\boldsymbol{r}^{m}|\boldsymbol{z}_{C}^{m})$. Minimizing $\mathcal{R}_{r}$ will encourage CRAB to infer the same rules on different context images of an RPM. The \textbf{target regularizer} $\mathcal{R}_{t}$ measures the distance between the concepts encoded from the target images and those predicted from the context. The encoded concepts can provide more accurate information about target images to guide the training of target predictors. The \textbf{context regularizer} considers the KL divergence between $q(\boldsymbol{z}_{c}^{a}|\boldsymbol{x}_{c})$ and $p(\boldsymbol{z}_{c}^{m}|\boldsymbol{x}_{c})$. Since CRAB shares the encoder in the generative and inference processes to reduce parameters, we have $q(\boldsymbol{z}_{c}^{m}|\boldsymbol{x}_{c})=p(\boldsymbol{z}_{c}^{m}|\boldsymbol{x}_{c})$ and $\mathcal{R}_{c}=0$. Thus $\mathcal{R}_{c}$ is removed from the ELBO to save computational resources. According to \cite{higgins2016beta}, we introduce hyperparameters $\beta_{r}$ and $\beta_{t}$ to control the importance of $\mathcal{R}_{r}$ and $\mathcal{R}_{t}$ in training. $\mathcal{L}$ is approximated with a Monte Carlo estimator by sampling from the variational distribution $q(\boldsymbol{h}|\boldsymbol{x})$ (see details in Appendix A).

\subsection{Module Instantiation}

According to Figure \ref{fig:svcl-process}, the probabilities in Equation \ref{eq:svcl-gen} are instantiated as
\begin{equation}
  \begin{aligned}
    p\left(\boldsymbol{z}^{m}_{c}|\boldsymbol{x}_{c}\right) &= \mathcal{N} \left(\boldsymbol{\mu}^{\text{con}}_{c,m}, \sigma_{z}^{2}\boldsymbol{I}\right), \\
    p\left(\boldsymbol{r}^{m}|\boldsymbol{z}^{m}_{C}\right) &= \mathcal{N} \left(\boldsymbol{\mu}^{\text{rul}}_{m}, \text{diag}((\boldsymbol{\sigma}^{\text{rul}}_{m})^{2})\right), \\
    p\left(\boldsymbol{z}^{m}_{T}|\boldsymbol{r}^{m},\boldsymbol{z}^{m}_{C}\right) &= \mathcal{N} \left(\boldsymbol{\mu}^{\text{con}}_{t,m}, \sigma_{z}^{2}\boldsymbol{I}\right), \\
    p\left(\boldsymbol{x}_{t}|\boldsymbol{z}_{t}\right) &= \mathcal{N}\left(\boldsymbol{\mu}^{\text{rec}}_{t}, \sigma_{x}^{2}\boldsymbol{I}\right).
  \end{aligned}
\end{equation}
The concept $\boldsymbol{z}^{m}_{c}$ is Gaussian distributed where the mean is produced via $\boldsymbol{\mu}^{\text{con}}_{c,:} = f_{\text{enc}}(\boldsymbol{x}_{c})$ and the standard deviation $\sigma_{z}$ is a hyperparameter to control the noise in sampling. Based on the concepts of context images, the concept-specific rule parsers and target predictors will take abstract reasoning. The parameters of the rule latent variable are computed by $\boldsymbol{\mu}^{\text{rul}}_{m},\boldsymbol{\sigma}^{\text{rul}}_{m} = f^{m}_{\text{rule}}(\boldsymbol{z}^{m}_{C})$. In the following target prediction process, the target predictors will output the mean of target concepts, i.e., $\boldsymbol{\mu}^{\text{con}}_{T,m} = f^{m}_{\text{target}}(\boldsymbol{z}^{m}_{C},\boldsymbol{r}^{m})$.
Finally, the decoder predicts the mean pixel values of target images by $\boldsymbol{\mu}^{\text{rec}}_{t} = f_{\text{dec}}(\boldsymbol{z}_{t})$, and $\sigma_x$ is the hyperparameter that controls the noise on reconstructed images.

By sharing the encoder and rule parsers in the generative and inference processes, the subterms in Equation \ref{eq:svcl-inf} become
\begin{equation}
  \begin{aligned}
    q\left(\boldsymbol{z}^{m}_{n}|\boldsymbol{x}_{n}\right) &= \mathcal{N}\left(\boldsymbol{\tilde{\mu}}^{\text{con}}_{n,m}, \sigma^{2}_{z}\boldsymbol{I}\right), \\
    q\left(\boldsymbol{r}^{m}|\boldsymbol{z}^{m}\right) &= \mathcal{N}\left(\boldsymbol{\tilde{\mu}}^{\text{rul}}_{m}, \text{diag}((\boldsymbol{\tilde{\sigma}}^{\text{rul}}_{m})^{2})\right).
  \end{aligned}
\end{equation}
where $\boldsymbol{\tilde{\mu}}^{\text{con}}_{n,:} = f_{\text{enc}}(\boldsymbol{x}_{n})$ and $\boldsymbol{\tilde{\mu}}^{\text{rul}}_{m}, \boldsymbol{\tilde{\sigma}}^{\text{rul}}_{m} = f^{m}_{\text{rule}}(\boldsymbol{z}^{m})$. Parameter sharing requires the rule parsers to infer rules on the subset $\boldsymbol{z}^{m}_{C} \subseteq \boldsymbol{z}^{m}$ to model the distributions $p(\boldsymbol{r}^{m}|\boldsymbol{z}^{m}_{C})$ and $q(\boldsymbol{r}^{m}|\boldsymbol{z}^{m})$. After instantiating the generative and inference processes, the model can adjust the parameters by maximizing Equation \ref{eq:svcl-elbo-regularizers}. With the fixed standard deviation, $\mathcal{R}_{t}$ can be regarded as the square error between the mean of Gaussians. And in $\mathcal{R}_{r}$, the KL divergences have closed-form solutions. We use the reparameterization trick to sample Gaussian-distributed random variables so that CRAB can be trained end-to-end with gradient descent optimizers.

\textbf{Rule Parser}. CRAB discovers the rule on $\boldsymbol{z}^{m}_{C} \subseteq \boldsymbol{z}^{m}$ through the concept-specific rule parser $f^{m}_{\text{rule}}$. The left panel in Figure \ref{fig:svcl-arch} is the overview of $f^{m}_{\text{rule}}$. CRAB takes the strategy of zero-padding to handle unknown target concepts at different positions, that is, we initialize the positions of target concepts with zero vectors to construct $\boldsymbol{Z}^{m}$. Then CRAB concatenates the concepts of every two images to form concept pairs $\{(\boldsymbol{Z}^{m}_{i}, \boldsymbol{Z}^{m}_{j}); i,j = 1,...,9\}$. The representations of concept pairs are extracted by a neural network $f^{m}_{\text{pair}}$:
\begin{equation}
  \begin{aligned}
      \boldsymbol{o}^{m}_{i,j} = g_{i,j} \cdot f^{m}_{\text{pair}} \left(\boldsymbol{Z}^{m}_{i}, \boldsymbol{Z}^{m}_{j}\right), \quad i,j=1,...,9.
  \end{aligned}
\end{equation}
To mask out $\boldsymbol{o}^{m}_{i,j}$ involving target positions, we set the gate variable $g_{i,j}=0$ when $i \in T$ or $j \in T$ and otherwise $g_{i,j}=1$. Since the pair representations of target images are set to zeros, $\boldsymbol{o}^{m}_{i,j}$ only contains the information about context images. To predict the rule on the entire matrix, CRAB introduces a learnable network $f^{m}_{\text{relation}}$ to convert the pair representations into the parameters of $\boldsymbol{r}^{m}$:
\begin{equation}
  \begin{aligned}
    \boldsymbol{\mu}^{\text{rul}}_{m}, \boldsymbol{\sigma}^{\text{rul}}_{m} = f^{m}_{\text{relation}} \left( s \cdot \boldsymbol{o}^{m}_{1:9,1:9} \right).
  \end{aligned}
\end{equation}
We set $s=1/81$ to reduce the input values and backpropagated gradients. In the rule parsing process, the rule parser $f^{m}_{\text{rule}}$ consists of the networks $f^{m}_{\text{pair}}$ and $f^{m}_{\text{relation}}$.

\textbf{Target Predictor}. To predict the concept of target images, the concept-specific target predictor estimates the conditional probability $p(\boldsymbol{z}^{m}_{T}|\boldsymbol{z}^{m}_{C},\boldsymbol{r}^{m})$. The right panel in Figure \ref{fig:svcl-arch} illustrates the architecture of target predictors. First, we concatenate $\boldsymbol{r}^{m}$ to each position of $\boldsymbol{Z}^{m}$ to create a 3$\times$3 input matrix where $\boldsymbol{Z}^{m}$ and $\boldsymbol{r}^{m}$ provide the context information of each image and the knowledge of underlying rules. The zero-padded vectors in the input matrix are iteratively updated to form the concept representations of target images. As Figure \ref{fig:svcl-arch} shows, the iterative process is realized by $f^{m}_{\text{target}}$ that contains four convolutional layers with the 3$\times$3 kernels and stride 1. The matrix is padded into 5$\times$5 with zeros before each convolutional layer to keep the output shape. After four iterations, the target predictor outputs the concept representations at the target positions $\boldsymbol{\mu}^{\text{con}}_{T,m}$, which is the mean of target concepts.

\subsection{Global Rule Abstraction}

CRAB can discover global rules and form the prior knowledge of rules shared on the entire dataset. The prior knowledge of rules on concept $m$ is captured with a mixture of $K$ Gaussians $p(\boldsymbol{r}^{m})=\text{GMM}(\boldsymbol{w}^{m}_{1:K}$, $\boldsymbol{\Sigma}^{m}_{1:K}$, $\boldsymbol{\mu}^{m}_{1:K})$ where each component indicates a type of global rule. CRAB regards $\boldsymbol{w}^{m}_{1:K}$, $\boldsymbol{\Sigma}^{m}_{1:K}$, and $\boldsymbol{\mu}^{m}_{1:K}$ as learnable parameters to enable the update of prior knowledge in the learning process.

\begin{algorithm}
  \caption{CRAB Iteration}\label{alg:svcl-iteration}
  \begin{algorithmic}
    \Require prior $p$, posterior $q$, batches of samples $\{\mathcal{B}_{1},...,\mathcal{B}_{U}\}$
    \For {$m \gets 1,M$}
      \State $\mathcal{R}^{m}_{1} = \{\}$
      \For {$\boldsymbol{x} \in \mathcal{B}_{1}$}
        \State $\tilde{\boldsymbol{z}}^{m} \sim q(\boldsymbol{z}^{m}|\boldsymbol{x})$
        \State $\tilde{\boldsymbol{r}}^{m} \sim q(\boldsymbol{r}^{m}|\tilde{\boldsymbol{z}}^{m})$
        \State $\mathcal{R}^{m}_{1} \gets \mathcal{R}^{m}_{1} \cup \{\tilde{\boldsymbol{r}}^{m}\}$
      \EndFor
      \State $\boldsymbol{w}^{m}_{1:K}$, $\boldsymbol{\Sigma}^{m}_{1:K}$, $\boldsymbol{\mu}^{m}_{1:K} = \text{GMMLeaner}(\mathcal{R}^{m}_{1})$
      \State $p(\boldsymbol{r}^{m}) \gets \text{GMM}(\boldsymbol{w}^{m}_{1:K}$, $\boldsymbol{\Sigma}^{m}_{1:K}$, $\boldsymbol{\mu}^{m}_{1:K})$
    \EndFor
    \For {$u \gets 2,U$}
      \State Compute $\mathcal{L}_{r}$ on $\mathcal{B}_{u}$ via Equation \ref{eq:svcl-elbo-lr}
      \State Compute $\mathcal{R}_{r}$, and $\mathcal{R}_{t}$ on $\mathcal{B}_{u}$ via Equation \ref{eq:svcl-elbo-regularizers}
      \State Compute $\mathcal{R}_{\text{cls}}$ on $\mathcal{B}_{u}$ via Equation \ref{eq:regularizer-prior}
      \State $p,q \gets \underset{p,q}{\argmax} \mathcal{L}_{r} - \beta_{r}\mathcal{R}_{r} - \beta_{t}\mathcal{R}_{t} - \beta_{\text{cls}}\mathcal{R}_{\text{cls}}$
    \EndFor
    \Ensure $p$, $q$
  \end{algorithmic}
\end{algorithm}

At the beginning of training, CRAB obtains the ability of rule parsing by maximizing the ELBO $\mathcal{L}$. Then CRAB abstracts global rules iteratively where a single iteration described in Algorithm 1 includes the knowledge update and knowledge-guided rule parsing stages. In knowledge update, CRAB acquires the means, covariances, and weights by learning a Gaussian mixture model on the rules $\mathcal{R}_{1}$ parsed from the batch of samples $\mathcal{B}_{1}$ to renovate the parameters of $p(\boldsymbol{r}^{m})$. In knowledge-guided rule parsing, CRAB fixes $p(\boldsymbol{r}^{m})$ and guides the inference of rules through the aggregated posterior $q(\boldsymbol{r}^{m})=\mathbb{E}_{q(\boldsymbol{z}^{m},\boldsymbol{x})}[q(\boldsymbol{r}^{m}|\boldsymbol{z}^{m})]$ \cite{makhzani2015adversarial}, which can be approximated on a batch of samples. CRAB adds $\mathcal{R}_{\text{cls}} = \sum_{m=1}^{M} D_{KL}(q(\boldsymbol{r}^{m})||p(\boldsymbol{r}^{m}))$ to the ELBO as a regularizer of the rule parsing. That is, minimizing the KL divergence between $q(\boldsymbol{r}^{m})$ and $p(\boldsymbol{r}^{m})$ will guide the distribution of the parsed rules close to the prior $p(\boldsymbol{r}^{m})$. Since $\mathcal{R}_{\text{cls}}$ has no closed-form solution, it is estimated through
\begin{equation}
  \begin{split}
    \mathcal{R}_{\text{cls}} = \sum_{m=1}^{M} \left( \mathbb{E}_{q(\boldsymbol{r}^{m})} \left[\log q(\boldsymbol{r}^{m})\right] - \mathbb{E}_{q(\boldsymbol{r}^{m})} \left[\log p(\boldsymbol{r}^{m})\right] \right).
  \end{split}
  \label{eq:regularizer-prior}
\end{equation}
The first term $\mathbb{E}_{q(\boldsymbol{r}^{m})}[\log q(\boldsymbol{r}^{m})]$ is estimated by Minibatch Weighted Sampling \cite{chen2018isolating}, and the log-likelihood in the second term is approximated with samples from $q(\boldsymbol{r}^{m})$ (see details in Appendix A). We also introduce a hyperparameter $\beta_{\text{cls}}$ to control the importance of $\mathcal{R}_{\text{cls}}$. The iterative learning process of CRAB makes it possible to abstract global rules as prior knowledge, which is updated in training to guide the learning of concept-specific rules.

\section{Experiments}

\begin{figure}[t]
  \centering
  \includegraphics[width=0.45\textwidth]{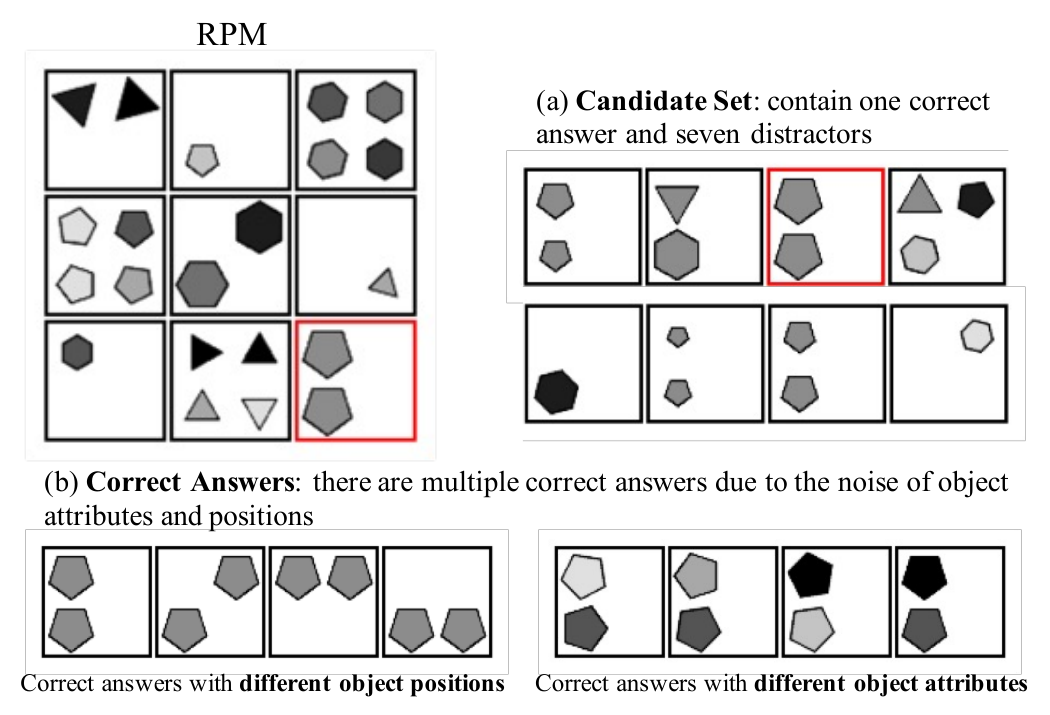}
  \caption{\textbf{The noise on RAVEN and I-RAVEN}. Panel (a) is the candidate set with one correct answer and seven distractors. Panel (b) illustrates that the RPM can have multiple correct answers due to the noise of object attributes and positions.}
  \label{fig:exp-dataset}
\end{figure}

This section introduces the experiment configurations (e.g., the datasets, compared models, and evaluation metrics) and conducts experiments to evaluate the abstract reasoning ability of the models. (1) The results of bottom-right and arbitrary-position answer generation are provided in Sections \ref{subsec:bottom-right-answer-generation} and \ref{subsec:arbitrary-position-answer-generation}, respectively. (2) In Section \ref{subsec:concept-decomposition}, we illustrate the interpretability of concepts by interpolating the concept representations. (3) Section \ref{subsec:interpretable-selections} explains how CRAB selects and excludes candidates through concept-wise distances between candidates and generation results. (4) In Section \ref{subsec:global-rule-abstraction}, we probe CRAB's global rule abstraction ability by visualizing the rule distributions at different training epochs and the RPMs sampled from mixture components.

\textbf{Datasets.} We use seven image configurations of RAVEN \cite{zhang2019raven} and I-RAVEN \cite{hu2021stratified} in the experiments to evaluate CRAB and the compared models. As Figure \ref{fig:exp-dataset} shows, the configurations with object grids (O-IG, 2$\times$2Grid, and 3$\times$3Grid) introduce noise to object attributes and positions to increase the difficulty of RPMs \cite{zhang2019raven}. For generative solvers, the noise makes the answer of RPMs not unique and brings obstacles in training especially when the auxiliary information is not provided. In the experiments, we remove the noise of object attributes from the original O-IG, 2$\times$2Grid, and 3$\times$3Grid configurations to reduce the number of correct answers to evaluate the abstract reasoning ability.

\textbf{Compared Models and Baselines.} We compare CRAB with the bottom-right generative solvers PrAE \cite{zhang2021abstract}, ALANS \cite{zhang2021learning}, and the method proposed by Niv Pekar et al. \cite{pekar2020generating} (called GCA for convenience). They are trained with auxiliary supervision to generate answers at the bottom-right of RPMs. We also compare CRAB with the bottom-right generative solver LoGe \cite{yu2021abstract} trained without auxiliary supervision. LoGe cannot handle the rule on object number and position to solve RPMs having object grids (e.g., O-IG, 2$\times$2Grid, and 3$\times$3Grid). To evaluate the answer generation ability at non-bottom-right positions, we introduce arbitrary-position solvers Transformer \cite{vaswani2017attention}, ANP \cite{kim2019attentive}, LGPP \cite{shi2021raven}, and CLAP \cite{shi2022compositional} (we use NP to instantiate CLAP, i.e., CLAP-NP) that can predict arbitrary-position answers without auxiliary supervision. Therefore, they are taken as the baseline models to illustrate CRAB's abstract reasoning ability in the more challenging arbitrary-position generation task.

\textbf{Metrics.} We use \textit{Selection Accuracy} (SA) on candidate sets to quantitatively evaluate the answer generation ability. The generative solvers choose the candidate closest to the prediction result. PrAE and ALANS use Jensen-Shannon Divergence as the distance measure in the representation space. GCA can select answers according to L2 distances on image pixels (GCA-I), Euclidean distances on representations (GCA-R), and outputs of the scoring network (GCA-C). CRAB and the baselines select answers through L2 distances on latent concepts or representations. Since RAVEN and I-RAVEN provide only candidate sets for bottom-right images, we generate candidate sets for non-bottom-right images with an in-batch sampling strategy. The candidate sets are generated by concatenating the correct answer of an RPM with the $N_{C}-1$ distractors from the same position of other in-batch samples. Using the generated candidate sets, we can compute selection accuracies at random positions of an RPM. The \textit{Global Selection Accuracy} (GSA) averages SAs at randomly selected positions of the matrix to estimate the ability of arbitrary-position answer generation.

\begin{table*}[t]
  \caption{\textbf{The accuracy (\%) of selecting bottom-right answers on RAVEN/I-RAVEN}. The second column (Aux) indicates the auxiliary supervision used in the training process, i.e., the distractors in candidate sets (D) and rule annotations (R).}
  \label{tab:sa-bottom-right}
  \centering
  \begin{tabular}{c|c|c|ccccccc}
  
  \hline
  Selective Solvers & Aux & Average & Center & L-R & U-D & O-IC & O-IG & 2$\times$2Grid & 3$\times$3Grid \\
  \hline
  LSTM \cite{barrett2018measuring} & D & 13.2/12.9 & 14.2/12.3 & 11.7/13.4 & 12.9/12.2 & 12.2/13.2 & 13.1/12.8 & 14.3/13.6 & 13.8/13.1 \\
  CNN \cite{barrett2018measuring} & D & 16.1/12.9 & 15.6/12.8 & 16.1/13.5 & 16.2/12.0 & 16.2/12.1 & 18.3/14.3 & 13.7/13.6 & 16.6/12.3 \\
  ResNet-50 \cite{barrett2018measuring} & D & 21.2/12.9 & 18.7/14.0 & 24.0/13.6 & 21.9/12.9 & 23.1/12.2 & 21.0/13.0 & 20.2/12.7 & 19.7/11.7 \\
  LSTM+DRT \cite{zhang2019raven} & D+R & 14.1/13.1 & 14.1/13.4 & 11.9/13.9 & 12.8/12.1 & 13.6/14.4 & 13.1/13.2 & 16.8/12.9 & 16.1/11.8 \\
  CNN+DRT \cite{zhang2019raven} & D+R & 14.0/13.0 & 13.5/12.6 & 11.4/13.5 & 16.5/11.7 & 12.2/15.1 & 15.1/12.5 & 14.8/12.9 & 14.8/12.6 \\
  ResNet+DRT \cite{zhang2019raven} & D+R & 36.2/12.8 & 33.3/13.2 & 46.9/13.4 & 41.9/12.1 & 41.6/12.1 & 32.3/13.3 & 30.2/12.4 & 26.9/12.8 \\
  WReN \cite{barrett2018measuring} & D & 22.9/38.1 & 14.0/46.9 & \,\,\,8.6/52.1 & \,\,\,9.3/49.3 & \,\,\,8.1/55.2 & 23.0/35.9 & 44.2/13.9 & 53.3/13.4 \\
  SRAN \cite{hu2021stratified} & D & 56.1/61.0 & 75.2/89.6 & 31.0/67.6 & 33.2/70.9 & 39.3/75.7 & 68.0/52.2 & 66.9/38.6 & 79.3/32.2 \\
  LEN \cite{zheng2019abstract} & D & 72.4/15.0 & 69.3/15.3 & 74.5/14.6 & 74.2/15.5 & 72.8/12.8 & 77.6/15.7 & 65.0/15.1 & 73.5/16.1 \\
  CoPINet \cite{zhang2019learning} & D & \textbf{95.6}/15.2 & 99.4/17.1 & \textbf{99.5}/13.9 & 99.5/14.5 & 98.7/13.2 & \textbf{94.9}/15.2 & 88.1/16.3 & \textbf{89.3}/16.4 \\
  SCL \cite{wu2020scattering} & D & 95.1/\textbf{85.7} & \textbf{99.9}/\textbf{99.9} & 99.1/\textbf{99.7} & \textbf{99.9}/\textbf{99.9} & \textbf{99.8}/\textbf{99.2} & 92.0/\textbf{80.6} & \textbf{98.0}/\textbf{78.2} & 77.0/\textbf{42.5} \\
  \hline
  \hline
  Generative Solvers & Aux & Average & Center & L-R & U-D & O-IC & O-IG & 2$\times$2Grid & 3$\times$3Grid \\
  \hline
  GCA-I \cite{pekar2020generating} & D+R & 14.9/29.5 & 14.0/34.4 & \,\,\,7.9/28.3 & \,\,\,7.5/26.7 & 13.4/43.8 & 21.2/38.0 & 19.5/18.3 & 20.6/17.2 \\
  GCA-R \cite{pekar2020generating} & D+R & 17.0/33.3 & 16.6/39.0 & \,\,\,9.4/32.3 & \,\,\,6.9/27.3 & 17.3/51.7 & 20.7/39.9 & 21.9/22.3 & 25.9/20.9 \\
  GCA-C \cite{pekar2020generating} & D+R & 42.1/45.8 & 37.3/56.8 & 26.4/57.4 & 21.5/45.7 & 30.2/60.3 & 53.8/41.7 & 58.8/31.8 & 67.0/26.9 \\
  ALANS \cite{zhang2021learning} & D+R & 54.9/64.5 & 45.4/61.2 & 45.7/67.7 & 44.0/65.0 & 38.0/64.9 & 47.1/55.0 & 77.6/67.9 & 86.4/69.8 \\
  PrAE \footnotemark \cite{zhang2021abstract} & D+R & 83.3/86.6 & 97.2/98.6 & \textbf{96.4}/\textbf{98.3} & \textbf{96.7}/\textbf{98.0} & \textbf{95.8}/98.2 & 79.4/80.0 & 84.0/83.1 & 33.6/50.2 \\
  \hline
  GCA-I \cite{pekar2020generating} & D & \,\,\,9.8/22.1 & 10.6/21.8 & \,\,\,5.4/21.3 & \,\,\,4.9/16.2 & 10.1/28.5 & 14.1/33.3 & 12.1/16.4 & 11.2/17.0 \\
  GCA-R \cite{pekar2020generating} & D & 12.8/26.7 & 11.6/26.7 & \,\,\,7.7/24.3 & \,\,\,5.0/20.8 & 15.0/37.9 & 18.2/36.6 & 16.8/22.0 & 15.2/18.8 \\
  GCA-C \cite{pekar2020generating} & D & 37.3/39.5 & 32.8/48.2 & 19.5/46.8 & 15.5/37.8 & 26.5/50.5 & 45.2/40.9 & 54.7/29.2 & 67.1/23.2 \\
  ALANS \cite{zhang2021learning} & D &  50.1/60.8 & 32.2/51.9 & 45.0/68.8 & 40.0/60.0 & 36.0/58.3 & 44.7/52.8 & 70.5/65.1 & 82.6/68.4 \\
  PrAE \cite{zhang2021abstract} & D & 13.6/24.7 & 14.5/22.6 & \,\,\,7.1/21.2 & 11.1/26.5 & \,\,\,7.1/16.9 & 10.5/25.6 & 22.7/29.4 & 22.1/30.5 \\
  \hline
  LoGe \footnotemark \cite{yu2021abstract} & - & \,\,\,\,--\,\, /62.9 & \,\,\,\,--\,\, /87.5 & \,\,\,\,--\,\, /51.7 & \,\,\,\,--\,\, /64.0 & \,\,\,\,--\,\, /48.5 & \,\,\,\,--\,\, / \,\,--\,\,\,\, & \,\,\,\,--\,\, / \,\,--\,\,\,\, & \,\,\,\,--\,\, / \,\,--\,\,\,\, \\
  LGPP \cite{shi2021raven} & - & \,\,\,4.9/16.1 & \,\,\,9.2/20.1 & \,\,\,4.7/18.9 & \,\,\,5.2/21.2 & \,\,\,4.0/13.9 & \,\,\,3.4/12.3 & \,\,\,4.1/13.0 & \,\,\,4.0/13.1 \\
  ANP \cite{kim2019attentive} & - & 10.9/27.5 & \,\,\,9.8/47.4 & \,\,\,4.1/20.3 & \,\,\,3.5/20.7 & \,\,\,5.4/38.2 & 31.5/34.0 & 10.0/15.6 & 12.0/16.3 \\ 
  CLAP-NP \cite{shi2022compositional} & - & 17.3/35.9 & 30.4/42.9 & 13.4/35.1 & 12.2/32.1 & 16.4/37.5 & 14.4/31.7 & 22.5/39.1 & 12.1/32.9 \\
  Transformer \cite{vaswani2017attention} & - & 59.8/73.5 & \textbf{98.4}/\textbf{99.2} & 67.0/91.1 & 60.9/86.6 & 14.5/69.9 & 70.6/57.9 & 73.3/73.0 & 34.2/37.0 \\
  CRAB & - & \textbf{94.7}/\textbf{95.9} & 96.6/97.1 & 93.9/96.7 & 95.3/97.5 & 94.6/\textbf{98.9} & \textbf{96.6}/\textbf{96.9} & \textbf{95.8}/\textbf{93.3} & \textbf{90.2}/\textbf{91.1} \\
  \hline
  \hline
  \textit{Human} \cite{zhang2019raven} & - & \textit{84.4} & \textit{95.5} & \textit{81.8} & \textit{79.6} & \textit{86.4} & \textit{81.8} & \textit{86.4} & \textit{81.8} \\
  \hline
  \end{tabular}
  \end{table*}

  In the experiments, the models are trained and evaluated independently in each image configuration of RAVEN. We acquire the selection accuracies on I-RAVEN by directly evaluating the model trained on RAVEN to analyze the influence of candidate sets in model performance. Due to the page limitation, we put the detailed introduction to datasets in Appendix B, the hyperparameter choice and model implementation in Appendix C, and the additional experimental results in Appendix D.

\subsection{Bottom-Right Answer Generation}
\label{subsec:bottom-right-answer-generation}

\begin{figure*}[t]
  \centering
  \includegraphics[width=0.85\textwidth]{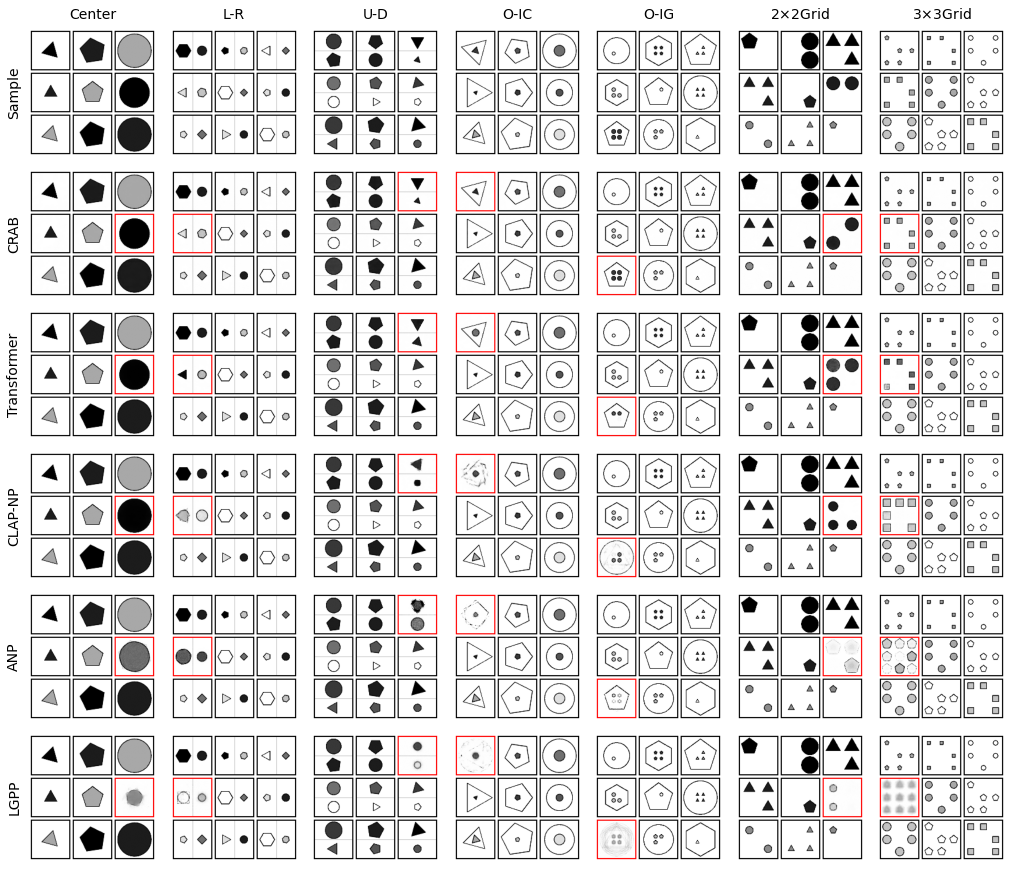}
  \caption{\textbf{Visualization results of arbitrary-position answer generation on RAVEN}. The original RPM samples are displayed in the first row, and the prediction results are highlighted with red boxes in each RPM.}
  \label{fig:exp-arb-gen-vis}
\end{figure*}

This experiment evaluates the ability of bottom-right answer generation. The SAs of CRAB and the compared models are provided in Table \ref{tab:sa-bottom-right}. For PrAE, ALANS, and GCA that require auxiliary supervision in training, we provide the accuracy of the models trained with distractors and rule annotations (D+R) and only rule annotations (D). Table \ref{tab:sa-bottom-right} shows that CRAB can generate more accurate answers than LoGe in the configurations without object grids. LoGe cannot solve RPMs from O-IG, 2$\times$2Grid, and 3$\times$3Grid since the reasoning process does not consider the rules on object number and position. In comparison, the reasoning process of CRAB can parse rules on concepts for RPMs with grids. CRAB achieves SAs comparable to PrAE, ALANS, and GCA in non-grid configurations and better scores in those with object grids. Without auxiliary supervision, CRAB outperforms the baseline models Transformer, ANP, LGPP, and CLAP-NP in most cases. These results verify that machine learning models can acquire the ability of bottom-right answer generation without auxiliary information.

\footnotetext[1]{In the original paper, PrAE is trained on 2$\times$2Grid samples and evaluated on all the image configurations to verify the cross-configuration generalization ability. Since training PrAE on 3$\times$3Grid samples requires too much computational resource, we follow the original configuration of PrAE to evaluate the performance on 3$\times$3Grid through the model trained on 2$\times$2Grid.}
\footnotetext[2]{LoGe only defines the logical reasoning process on the attributes of color, size, and type. Therefore, LoGe cannot handle RPMs from O-IG, 2$\times$2Grid, and 3$\times$3Grid. Since the official code of LoGe is not released, we only provide the selection accuracies on I-RAVEN as a reference.}

The SAs of RAVEN and I-RAVEN in Table \ref{tab:sa-bottom-right} can also illustrate the influence of distractors in training. The candidate sets of RAVEN have inductive biases that imply the correct answers, making it possible for models to find correct answers by observing only candidate sets. I-RAVEN adopts a more reasonable way to generate candidate sets to avoid the problem of shortcut learning in RAVEN. Therefore, the selection accuracies of selective solvers in Table \ref{tab:sa-bottom-right} are more likely to decline on I-RAVEN. The generative solvers choose answers based on prediction results, which lessens the probability of shortcut learning. Instead, they usually achieve higher accuracy on I-RAVEN since the method used to generate unbiased candidate sets may decrease the difficulty of excluding distractors. The average accuracy of RAVEN and I-RAVEN shows that selective solvers are more likely to fall into shortcut learning than generative solvers. CRAB has close accuracies on RAVEN and I-RAVEN, which indicates the robustness to the distribution of distractors.

\subsection{Arbitrary-Position Answer Generation}
\label{subsec:arbitrary-position-answer-generation} 

\begin{figure*}[t]
  \centering
  \includegraphics[width=0.85\textwidth]{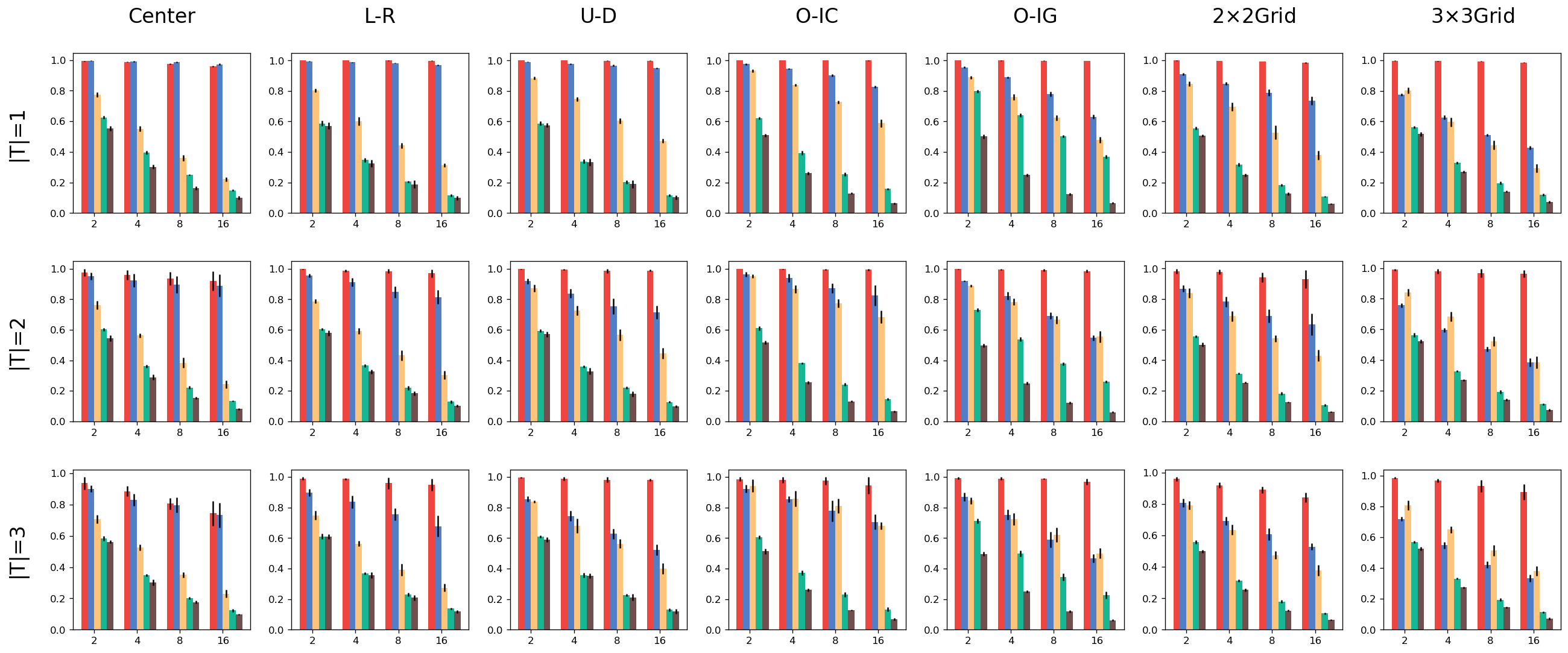}
  \caption{\textbf{Global Selection Accuracy (GSA) on RAVEN}. We display GSA scores of CRAB (red), Transformer (blue), CLAP-NP (orange), ANP (green), and LGPP (gray) in each plot where the y-axis is the GSA score, and the x-axis is the number of candidates. $|T|$ indicates the number of target images to predict.}
  \label{fig:exp-arb-gen-gsa}
\end{figure*}

This experiment evaluates the ability of arbitrary-position answer generation, where the models are trained to generate a random target image in the matrix and evaluated by generating answers at arbitrary or multiple positions. We test the generalization ability of the models by applying the acquired abstract reasoning ability to RPMs with multiple target images. Figure \ref{fig:exp-arb-gen-vis} visualizes the results of arbitrary-position answer generation on RAVEN ($|T|=1$). CRAB can generate accurate answers conforming to the underlying rules in all seven image configurations. Transformer can only generate high-quality answers in relatively simple configurations (e.g., Center and U-D). In complex configurations with inside-outside and grid layouts, Transformer may generate target images with incorrect attributes (e.g., wrong types of inner objects on O-IC). CLAP-NP tends to generate clear but incorrect target images. The predictions of ANP and LGPP are possibly blurred, leading to the relatively lower selection accuracy on RAVEN. The visualization results in Figure \ref{fig:exp-arb-gen-vis} illustrate CRAB's in-depth understanding of rules, which is the foundation of solving multiple-position answer generation problems without retraining.

To analyze the model performance under different configurations, Figure \ref{fig:exp-arb-gen-gsa} shows the quantitative results where the number of target images $|T| = 1,2,3$, and the candidate set size is chosen from $\{2,4,8,16\}$. CRAB outperforms the baseline models and has smaller accuracy declines when the size of candidate sets and the number of target images increase. A potential reason causing the performance difference is concept decomposition. Transformer achieves high GSA scores on Center, L-R, and U-D when $|T|=1$, and the accuracy declines obviously when $|T| \geq 2$. The decline in GSA scores indicates that Transformer has difficulty applying the acquired reasoning ability to RPMs with multiple target images. Transformer has performance comparable to CRAB in relatively simple configurations (e.g., Center) but struggles to generate correct answers in complex configurations (e.g., 3$\times$3Grid). ANP, CLAP-NP, and LGPP can hardly capture the underlying rules on the RPM from RAVEN-style datasets to generate answers in different configurations.

\subsection{Concept Learning}
\label{subsec:concept-decomposition}

\begin{figure}[t]
  \centering
  \includegraphics[width=0.4\textwidth]{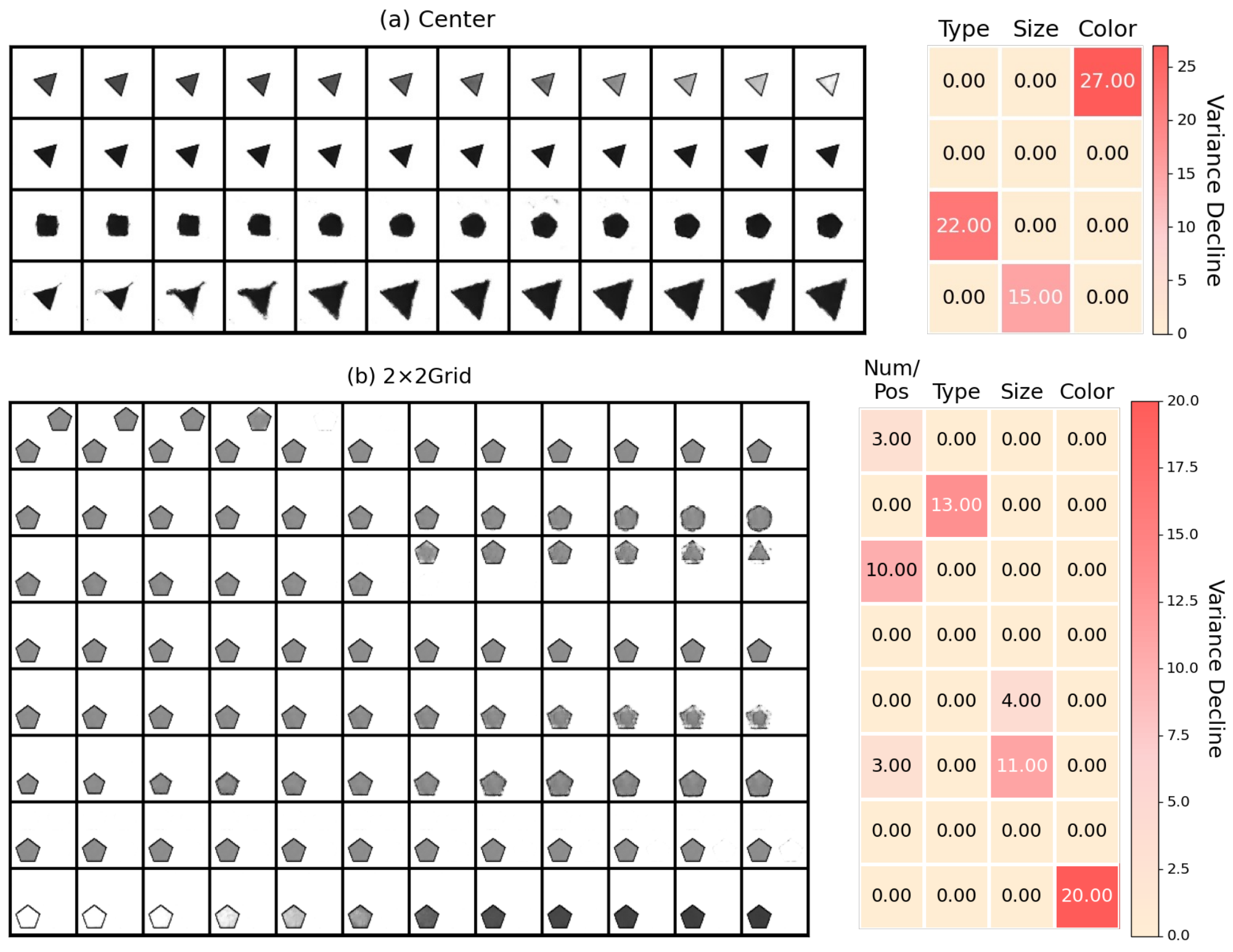}
  \caption{\textbf{Concept Learning on Center and 2$\times$2Grid}. The left panel of each plot is the interpolation results on different latent concepts. The right panel displays variance declines between concepts and real attributes. A high variance decline usually means that the concept encodes the attribute.}
  \label{fig:exp-concept-decompose}
\end{figure}

Figure \ref{fig:exp-concept-decompose} visualizes the learned concepts on O-IC and 2$\times$2Grid. We determine the correspondence between concepts and real attributes by computing variance declines (VDs). VD is calculated similarly to the FactorVAE score \cite{kim2018disentangling}. Assume that there are $V$ real attributes in the dataset, we randomly generate a batch of samples $\mathcal{B}$ and batches $\{\mathcal{B}_{v}\}_{v=1}^{V}$ by fixing each real attribute. Then we infer the concepts $\boldsymbol{z}_{\mathcal{B}}$ of samples in $\mathcal{B}$ and $\boldsymbol{z}_{\mathcal{B}_{v}}$ of $\mathcal{B}_{v}$. The VD between the concept $m$ and attribute $v$ is $\text{VD}_{m,v} = \text{Var}(\boldsymbol{z}^{m}_{\mathcal{B}_{v}}) / \text{Var}(\boldsymbol{z}^{m}_{\mathcal{B}})$ where $\text{Var}(\boldsymbol{z}) = \log |\boldsymbol{z}^{T}\boldsymbol{z}|$ gives the in-batch variance of concepts $\boldsymbol{z}^{m}_{\mathcal{B}_{v}}$ and $\boldsymbol{z}^{m}_{\mathcal{B}}$. We use multiple dimensions to represent a concept in CRAB and thus take the log determinants of the latent concepts learned from samples of a batch to include the relationship between dimensions. The concept-specific rules are automatically learned from data, and the way to decompose the composed rule into concept-specific rules is not unique (e.g., one can decompose the rule of \textit{Num/Pos} into the rule of \textit{Num} and the rule of \textit{Pos}). Therefore, we do not use VD scores to evaluate the quality of concept learning. We utilize VD to discover the correspondence between concepts and real attributes.

The interpolation results and VDs in Figure \ref{fig:exp-concept-decompose} reveal the interpretability of the learned concepts. Although the attribute of matrix images is discrete, CRAB can generate continuous interpolation results for some concepts. For example, in Figure \ref{fig:exp-concept-decompose}a, the grayscale of triangles decreases smoothly as we interpolate the latent space. Figure \ref{fig:exp-concept-decompose}b shows the results of concept learning on 2$\times$2Grid where CRAB encodes the attribute \textit{Num/Pos} (the number and position of objects in the grid) in multiple concepts. Although CRAB does not learn concepts according to real attributes, this way of decomposition can still explain the rules of RPMs. It is worth noting that when we set too many concepts, CRAB will automatically generate redundant concepts that do not encode any information (i.e., the 4th concept in Figure \ref{fig:exp-concept-decompose}b). The interpolation results on redundant concepts will stay the same, and the VDs will close to zeros.

\subsection{Interpretable Answer Selection}
\label{subsec:interpretable-selections}

\begin{figure*}[!t]
  \centering
  \includegraphics[width=0.8\textwidth]{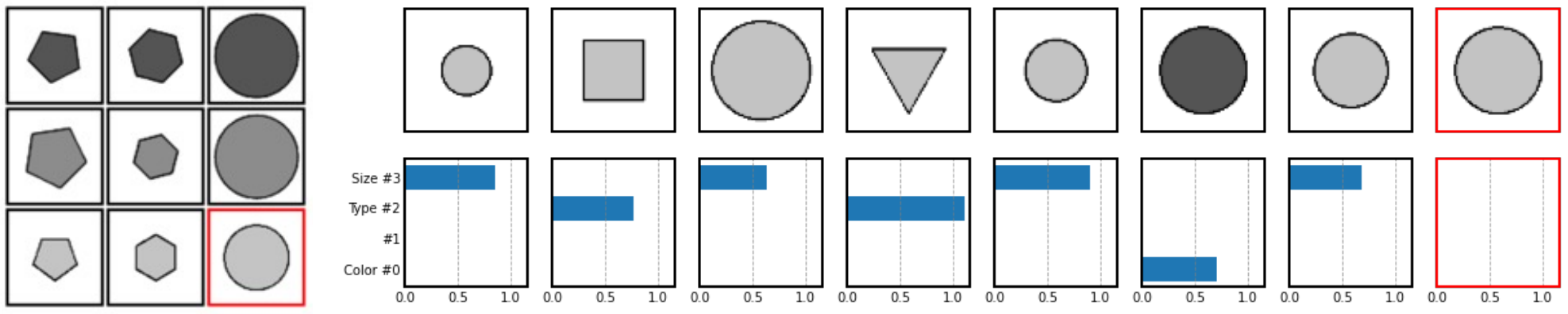}
  \caption{\textbf{The answer selection process of CRAB on Center}. The left panel provides the RPM and highlights the target image with the red box. The right panel contains the candidates (top) and concept-wise distances (bottom). We annotate the corresponding attribute for each concept according to the variance declines computed previously.}
  \label{fig:exp-interpretable-prediction}
\end{figure*}

\begin{figure*}[!t]
  \centering
  \includegraphics[width=0.9\textwidth]{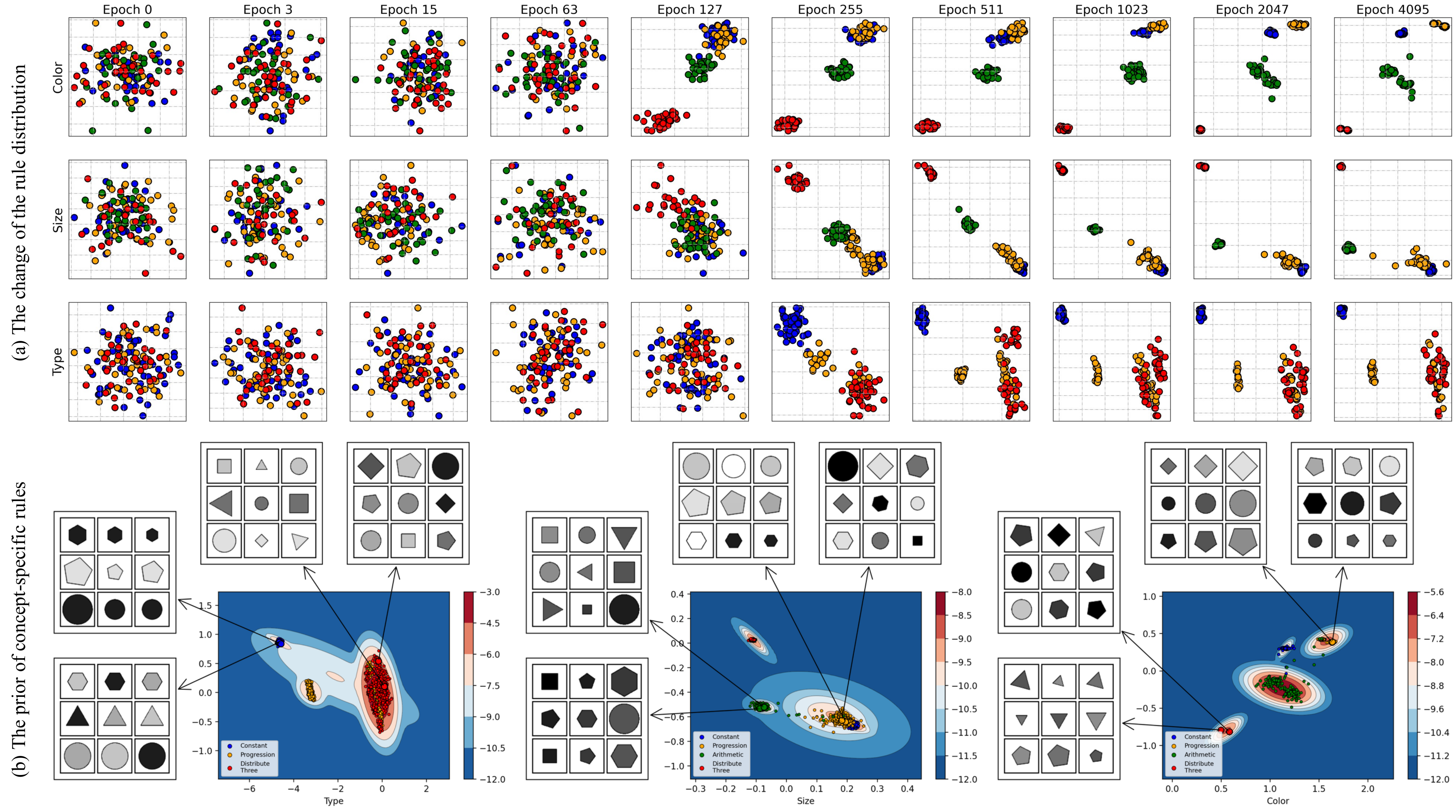}
  \caption{\textbf{Visualization results of global rule abstraction on Center}. Panel (a) displays the change of rule distributions on different concepts where the scatters of different colors represent the rule latent variables parsed from RPMs with different concept-specific rules. Panel (b) displays priors of concept-specific rules where the contours in the background indicate the log-likelihood $\log p(\boldsymbol{r}^{m})$. We visualize RPMs from the same cluster to illustrate the interpretability of the global rules.}
  \label{fig:exp-rule-cat}
\end{figure*}

To illustrate the process of answer selection in CRAB, we visualize the concept-wise distances between the prediction and candidate images to analyze the incorrect attributes of distractors. Figure \ref{fig:exp-interpretable-prediction} shows the concept-wise distances on samples of Center, where CRAB has no obvious prediction deviations on the correct answer. The third concept of the first candidate image differs from the prediction. We argue that this candidate has the wrong attribute \textit{Size} because the third concept encodes the attribute \textit{Size} according to the VDs. The concept-wise distances explain how the model excludes incorrect candidate images. The first candidate is excluded since the first concept (encoding the attribute \textit{Size}) breaks the rules. Similarly, the second candidate deviates from the prediction result on the second concept (encoding the attribute \textit{Type}). Sometimes, the magnitude of distances can indicate the degree of the rule violation. For example, the prediction error on \textit{Type} of the seventh candidate is smaller than that of the fifth candidate because the size of the seventh candidate is closer to the correct answer.

\subsection{Global Rule Abstraction}
\label{subsec:global-rule-abstraction}

This experiment evaluates the ability of global rule abstraction in CRAB qualitatively and quantitatively. The qualitative results of the configuration Center are displayed through the distribution of colored scatters in Figure \ref{fig:exp-rule-cat}. To estimate whether the model gathers the RPMs with the same concept-specific rule in a cluster, we put color on the scatters according to the rule annotations. The visualization results illustrate the global rule abstraction ability of CRAB on \textit{Type}, \textit{Size}, and \textit{Color}. Figure \ref{fig:exp-rule-cat}b reveals the ability to cluster RPMs with the same underlying rules together. We randomly select two scatters from the same cluster and visualize the corresponding RPM panels where the attribute \textit{Color} follows the rule \textit{Progression}. We also visualize two RPMs from the cluster in the bottom-left corner where the rule on \textit{Color} is \textit{Distribute Three}. In Figure \ref{fig:exp-rule-cat}b, it is observed that CRAB may assign the samples of the rule \textit{Progression} into different clusters. It is sometimes appropriate to allocate RPMs with the same concept-specific rule into different clusters because the rules provided in the annotations can be further decomposed into subrules (e.g., the rule \textit{Progression} consists of two rules of increase and decrease).

\begin{table}[!t]
  \caption{\textbf{ARI on each attribute}. We provide ARIs in terms of the real attributes (\textit{Num/Pos}, \textit{Type}, \textit{Size}, and \textit{Color}) and the average ARIs on different image configurations.}
  \label{tab:ari-attribute}
  \centering
  \begin{tabular}{c|c|cccc}
  \hline
  Configs & Average & Type & Size & Color & Num/Pos \\
  \hline
  Center & 0.7332 & 0.6018 & 0.6099 & \textbf{0.9878} & - \\
  L-R & 0.6963 & 0.6409 & \textbf{0.8540} & 0.5942 & - \\
  U-D & 0.7583 & 0.6225 & \textbf{0.8525} & 0.8001 & - \\
  O-IC & 0.3899 & 0.3026 & 0.1770 & \textbf{0.9903} & - \\
  O-IG & 0.3943 & 0.0766 & \textbf{0.9542} & 0.6059 & 0.2582 \\
  2$\times$2Grid & 0.4866 & 0.0039 & 0.6973 & \textbf{0.9932} & 0.2520 \\
  3$\times$3Grid & 0.5847 & 0.3916 & 0.7750 & \textbf{0.9872} & 0.1849 \\
  \hline
  Average & 0.5819 & 0.3771 & 0.7928 & \textbf{0.8512} & 0.2317 \\
  \hline
  \end{tabular}
\end{table}

In addition to the qualitative results, we quantitatively evaluate CRAB's rule abstraction ability on RAVEN. The performance of rule abstraction is evaluated by the Adjusted Rand Index (ARI) \cite{hubert1985comparing} that measures the clustering performance. The rule categories of each concept are given by assigning the rule latent variable to the prior mixture component with the maximum log-likelihood. We compute a $M \times V$ matrix of ARI scores where $\text{ARI}_{m,v}$ is computed with the rule annotations on the attribute $v$ and the rule categories predicted on the concept $m$. We obtain the optimal assignment by finding the concept with the maximum ARI for each attribute, and the final ARI is computed by averaging the ARIs on real attributes. Table \ref{tab:ari-attribute} shows the configuration-specific and average ARIs. From the quantitative results, we find that \textit{Color} and \textit{Size} have high ARI while abstracting \textit{Type} is more challenging. The performance of concept learning is highly related to the accuracy of global concept-specific rule abstraction. CRAB achieves high ARIs on Center since the concepts encode different attributes clearly. In contrast, the entangled concepts in O-IC may cause the incorrect assignment of attributes and pull down the ARI scores.

\section{Conclusion and Discussion}

We propose a deep latent variable model for Concept-changing Rule ABstraction (CRAB). CRAB is trained without auxiliary supervision to predict target images at arbitrary positions. CRAB parses concept-specific rules and discovers global rules shared on the dataset through an iterative learning process. CRAB achieves comparable or higher selection accuracies than the bottom-right generative solvers. In arbitrary-position answer generation, CRAB outperforms the baselines trained without auxiliary supervision. By conducting abstract reasoning through conditional generation, CRAB eliminates the requirement of auxiliary supervision and manifests the powerful arbitrary-position answer generation ability. Further experiments show that CRAB can automatically learn latent concepts and abstract the parsed concept-specific rules into global rules shared among RPMs. The interpretability of concept learning and global rule abstraction is an attempt to realize human-like abstract reasoning in machine intelligence. We discuss the limitations of CRAB in two aspects.
\begin{itemize}
  \item Although CRAB can generalize the answer generation ability to RPMs with multiple target images, the accuracy significantly declines as the number of target images increases. In the future, it is a challenge to predict targets with few context images or generate the entire RPMs, which requires the model to have a more in-depth and throughout understanding of underlying rules.
  \item As discussed in the experiment configurations, learning latent concepts and concept-specific rules from noisy data is another challenge. In this work, we remove the noise of object attributes to make the rules in RPMs from O-IG, 2$\times$2Grid, and 3$\times$3Grid easier to recognize. We need to explore methods in the future to handle data with more complex noise, e.g., the RPMs on PGM \cite{barrett2018measuring}.
\end{itemize}


%







\ifCLASSOPTIONcaptionsoff
  \newpage
\fi



\bibliographystyle{IEEEtran}
\bibliography{main}

%



%

\begin{IEEEbiography}[{\includegraphics[width=1in,height=1.25in,clip,keepaspectratio]{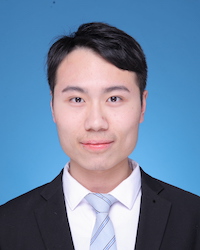}}]{Fan Shi}
  received the BS and MS degrees in computer science from Fudan University, China. He is currently a PhD candidate in the School of Computer Science, Fudan University, China. His research interests include abstract visual reasoning, machine learning, and deep generative models.
\end{IEEEbiography}

\begin{IEEEbiography}[{\includegraphics[width=1in,height=1.25in,clip,keepaspectratio]{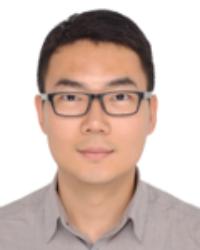}}]{Bin Li}
  received the PhD degree in computer science from Fudan University, Shanghai, China. He is an associate professor with the School of Computer Science, Fudan University, Shanghai, China. Before joining Fudan University, Shanghai, China, he was a lecturer with the University of Technology Sydney, Australia and a senior research scientist with Data61 (formerly NICTA), CSIRO, Australia. His current research interests include machine learning and visual intelligence, particularly in compositional scene representation, modeling and inference.
\end{IEEEbiography}

\begin{IEEEbiography}[{\includegraphics[width=1in,height=1.25in,clip,keepaspectratio]{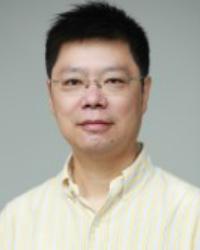}}]{Xiangyang Xue}
  received the BS, MS, and PhD degrees in communication engineering from Xidian University, Xian, China, in 1989, 1992, and 1995, respectively. He is currently a professor of computer science with Fudan University, Shanghai, China. His research interests include multimedia information processing and machine learning.
\end{IEEEbiography}







\appendices
\onecolumn

\section{Proofs and Derivations}

\subsection{Proof of the ELBO}

\begin{equation}
  \begin{aligned}
    \mathcal{L} &= \mathbb{E}_{q(\boldsymbol{h}|\boldsymbol{x})} \Big[ \log p\left(\boldsymbol{x}_{T}|\boldsymbol{h},\boldsymbol{x}_{C}\right) \Big] - \mathbb{E}_{q(\boldsymbol{h}|\boldsymbol{x})}\left[\log \frac{q\left(\boldsymbol{h}|\boldsymbol{x}\right)}{p\left(\boldsymbol{h}|\boldsymbol{x}_{C}\right)}\right] \\
    &= \sum_{t \in T} \mathbb{E}_{q(\boldsymbol{h}|\boldsymbol{x})}\Big[\log p\left(\boldsymbol{x}_{t} | \boldsymbol{z}_{t}\right)\Big] - \mathbb{E}_{q(\boldsymbol{h}|\boldsymbol{x})}\left[\log \prod_{m=1}^{M} \frac{q\left(\boldsymbol{r}^{m}|\boldsymbol{z}^{m}\right)\prod_{c \in C}q\left(\boldsymbol{z}_{c}^{m}|\boldsymbol{x}_{c}\right) \prod_{t \in T}q\left(\boldsymbol{z}_{t}^{m}|\boldsymbol{x}_{t}\right)}{p\left(\boldsymbol{r}^{m}|\boldsymbol{z}_{C}^{m}\right)p\left(\boldsymbol{z}_{T}^{m}|\boldsymbol{r}^{m},\boldsymbol{z}_{C}^{m}\right) \prod_{c \in C}p\left(\boldsymbol{z}_{c}^{m}|\boldsymbol{x}_{c}\right)} \right] \\
    &= \sum_{t \in T} \mathbb{E}_{q(\boldsymbol{h}|\boldsymbol{x})}\Big[\log p\left(\boldsymbol{x}_{t} | \boldsymbol{z}_{t}\right)\Big] - \sum_{m=1}^{M} \mathbb{E}_{q(\boldsymbol{h}|\boldsymbol{x})} \left[ \log \frac{q\left(\boldsymbol{r}^{m}|\boldsymbol{z}^{m}\right)}{p\left(\boldsymbol{r}^{m}|\boldsymbol{z}_{C}^{m}\right)} \right] \\
    & \quad \quad \quad - \sum_{m=1}^{M} \mathbb{E}_{q(\boldsymbol{h}|\boldsymbol{x})} \left[ \log \frac{q\left(\boldsymbol{z}_{T}^{m}|\boldsymbol{x}_{T}\right)}{p\left(\boldsymbol{z}_{T}^{m}|\boldsymbol{r}^{m},\boldsymbol{z}_{C}^{m}\right)}\right] - \underbrace{ \sum_{m=1}^{M} \sum_{c \in C} \mathbb{E}_{q(\boldsymbol{h}|\boldsymbol{x})} \left[ \log \frac{q\left(\boldsymbol{z}_{c}^{m}|\boldsymbol{x}_{c}\right)}{p\left(\boldsymbol{z}_{c}^{m}|\boldsymbol{x}_{c}\right)}\right] }_{\text{\small $q(\boldsymbol{z}_{c}^{m}|\boldsymbol{x}_{c}) = p(\boldsymbol{z}_{c}^{m}|\boldsymbol{x}_{c}) \Rightarrow \log \frac{q(\boldsymbol{z}_{c}^{m}|\boldsymbol{x}_{c})}{p(\boldsymbol{z}_{c}^{m}|\boldsymbol{x}_{c})} = 0$}} \\
    & = \underbrace{\sum_{t \in T} \mathbb{E}_{q(\boldsymbol{h}|\boldsymbol{x})}\Big[\log p\left(\boldsymbol{x}_{t} | \boldsymbol{z}_{t}\right)\Big]}_{\text{Reconstruction Term $\mathcal{L}_{r}$}} - \underbrace{ \sum_{m=1}^{M} \mathbb{E}_{q(\boldsymbol{h}|\boldsymbol{x})} \left[ \log \frac{q\left(\boldsymbol{r}^{m}|\boldsymbol{z}^{m}\right)}{p\left(\boldsymbol{r}^{m}|\boldsymbol{z}_{C}^{m}\right)} \right] }_{\text{Rule Regularizer $\mathcal{R}_{r}$}} - \underbrace{ \sum_{m=1}^{M} \mathbb{E}_{q(\boldsymbol{h}|\boldsymbol{x})} \left[ \log \frac{q\left(\boldsymbol{z}_{T}^{m}|\boldsymbol{x}_{T}\right)}{p\left(\boldsymbol{z}_{T}^{m}|\boldsymbol{r}^{m},\boldsymbol{z}_{C}^{m}\right)}\right] }_{\text{Target Regularizer $\mathcal{R}_{t}$}}
  \end{aligned}
\end{equation}

\subsection{Approximation of the ELBO}

The ELBO is approximated with the samples of the variational distribution $q(\boldsymbol{h}|\boldsymbol{x})$. That is, we first sample latent variables according to the inference process:
\begin{equation}
  \begin{aligned}
    \boldsymbol{\tilde{z}}^{m}_{n} &\sim q\left(\boldsymbol{z}^{m}_{n}|\boldsymbol{x}_{n}\right), &\quad n=1,...,9, \quad m=1,...,M, \\
    \boldsymbol{\tilde{r}}^{m} &\sim q\left(\boldsymbol{r}^{m}|\boldsymbol{\tilde{z}}^{m}\right), &\quad m=1,...,M.
  \end{aligned}
\end{equation}
According to \citeappendix{shi2022compositional_appendix}, the ELBO is approximated by
\begin{equation}
  \begin{aligned}
    \mathcal{L}_{r} &\approx \sum_{t \in T} \log p\left(\boldsymbol{x}_{t}|\tilde{\boldsymbol{z}}_{t}\right), \\
    \mathcal{R}_{r} &\approx \sum_{m=1}^{M} D_{KL}\left(q\left(\boldsymbol{r}^{m}|\tilde{\boldsymbol{z}}^{m}\right)\big\|p\left(\boldsymbol{r}^{m}|\tilde{\boldsymbol{z}}_{C}^{m}\right)\right), \\
    \mathcal{R}_{t} &\approx \sum_{m=1}^{M} D_{KL}\left(q\left(\boldsymbol{z}_{T}^{m}|\boldsymbol{x}_{T}\right)\big\|p\left(\boldsymbol{z}_{T}^{m}|\tilde{\boldsymbol{r}}^{m},\tilde{\boldsymbol{z}}_{C}^{m}\right)\right).
  \end{aligned}
\end{equation}
In $\mathcal{R}_{r}$ and $\mathcal{R}_{t}$, the KL divergences between Gaussians have closed-form solutions.

\subsection{Minibatch-Weighted Sampling of $\mathcal{R}_{\text{cls}}$}

With the aggregated posterior $q(\boldsymbol{r}^{m})=\mathbb{E}_{q(\boldsymbol{z}^{m},\boldsymbol{x})} [ q(\boldsymbol{r}^{m}|\boldsymbol{z}^{m}) ]$ and the prior of rules $p(\boldsymbol{r}^{m})=\text{GMM}(\boldsymbol{w}^{m}_{1:K}, \boldsymbol{\Sigma}^{m}_{1:K}, \boldsymbol{\mu}^{m}_{1:K})$, the process of rule parsing is guided by
\begin{equation}
  \begin{aligned}
    \mathcal{R}_{\text{cls}} = \sum_{m=1}^{M} D_{KL}(q(\boldsymbol{r}^{m})||p(\boldsymbol{r}^{m})) = \sum_{m=1}^{M} \left( \mathbb{E}_{q(\boldsymbol{r}^{m})} \left[\log q(\boldsymbol{r}^{m})\right] - \mathbb{E}_{q(\boldsymbol{r}^{m})} \left[\log p(\boldsymbol{r}^{m})\right] \right).
  \end{aligned}
\end{equation}
To avoid estimating $\mathbb{E}_{q(\boldsymbol{r}^{m})}[\log q(\boldsymbol{r}^{m})]$ on the whole dataset $\mathcal{D}$, we adopt Minibatch Weighted Sampling \citeappendix{chen2018isolating_appendix} to compute the expectation with the batch of $N_{\mathcal{B}}$ samples $\mathcal{B}=\{\boldsymbol{x}^{(b)}\}_{b=1}^{N_{\mathcal{B}}}$ and the concepts $\{\boldsymbol{z}^{m}(b)\}_{b=1}^{N_{\mathcal{B}}}$ where $\boldsymbol{z}^{m}(b) \sim q(\boldsymbol{z}^{m}|\boldsymbol{x}^{(b)})$:
\begin{equation}
  \begin{aligned}
    \mathbb{E}_{q(\boldsymbol{r}^{m})} \left[\log q(\boldsymbol{r}^{m})\right] \approx \frac{1}{N_{\mathcal{B}}} \sum_{i=1}^{N_{\mathcal{B}}} \left[ \log \frac{1}{N_{\mathcal{D}}N_{\mathcal{B}}} \sum_{j=1}^{N_{\mathcal{B}}} q\left(\boldsymbol{r}^{m}(i)|\boldsymbol{z}^{m}(j)\right) \right].
  \end{aligned}
\end{equation}
$\boldsymbol{r}^{m}(i)$ is the rule latent variable sampled from $q(\boldsymbol{r}^{m}|\boldsymbol{z}^{m}(i))$, and $N_{\mathcal{D}}$ is the size of the dataset $\mathcal{D}$. Then we adopt a Monte Carlo estimator to calculate $\mathbb{E}_{q(\boldsymbol{r}^{m})} \left[\log p(\boldsymbol{r}^{m})\right]$ with the rule latent variables sampled from the aggregated posterior $q(\boldsymbol{r}^{m})$. Since $q(\boldsymbol{r}^{m})=\mathbb{E}_{q(\boldsymbol{z}^{m},\boldsymbol{x})} [ q(\boldsymbol{r}^{m}|\boldsymbol{z}^{m}) ]$, we take $\{\boldsymbol{r}^{m}(b)\}_{b=1}^{N_{\mathcal{B}}}$ parsed from the sample batch $\mathcal{B}$ to compute the log-likelihoods in terms of the Gaussian mixture distributed prior:
\begin{equation}
  \begin{aligned}
    \mathbb{E}_{q(\boldsymbol{r}^{m})} \left[\log p(\boldsymbol{r}^{m})\right] \approx \frac{1}{N_{\mathcal{B}}} \sum_{b=1}^{N_{\mathcal{B}}} \log \sum_{k=1}^K w_{k}^{m} \mathcal{N}\left(\boldsymbol{r}^{m}(b)|\boldsymbol{\mu}_{k}^{m},\boldsymbol{\Sigma}_{k}^{m}\right).
  \end{aligned}
\end{equation}

\section{Datasets}

\begin{figure}[t]
  \centering
  \includegraphics[width=\textwidth]{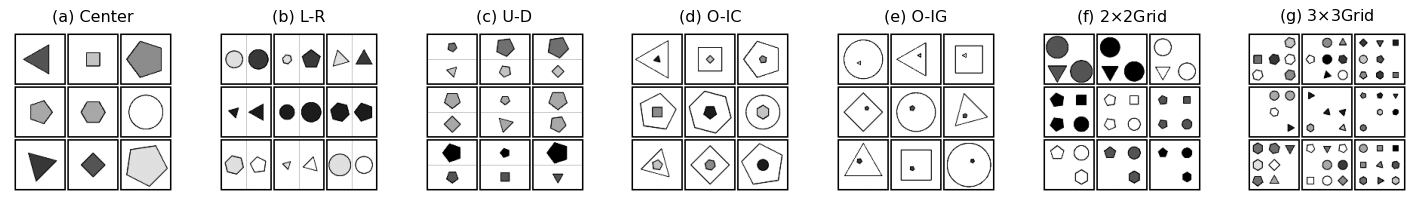}
  \caption{\textbf{The instances of seven image configurations}. RAVEN and I-RAVEN share the same image configurations.}
  \label{fig:datasets-instance}
\end{figure}

\begin{table}[t]
  \caption{The real attributes and their descriptions in each image configuration.}
  \label{tab:datasets-attribute}
  \centering
  \begin{tabular}{c|c|l}
  \hline
  Configurations & Attributes & Description \\
  \hline
  \multirow{3}{*}{Center} & Type & Type of the centric object \\
  & Size & Size of the centric object \\
  & Color & Color of the centric object \\ \hline
  \multirow{6}{*}{L-R} & Type Left & Type of the left object \\
  & Size Left & Size of the left object \\
  & Color Left & Color of the left object \\
  & Type Right & Type of the right object \\
  & Size Right & Size of the right object \\
  & Color Right & Color of the right object \\ \hline
  \multirow{6}{*}{U-D} & Type Up & Type of the top object \\
  & Size Up & Size of the top object \\
  & Color Up & Color of the top object \\
  & Type Down & Type of the bottom object \\
  & Size Down & Size of the bottom object \\
  & Color Down & Color of the bottom object \\ \hline
  \multirow{5}{*}{O-IC} & Type Out & Type of the outer object \\
  & Size Out & Size of the outer object \\
  & Type In & Type of the inner object \\
  & Size In & Size of the inner object \\
  & Color In & Color of the inner object \\ \hline
  \multirow{5}{*}{O-IG} & Type Out & Type of the outer object \\
  & Num/Pos In & Number and position of objects in the inner grid \\
  & Type In & Type of objects in the inner grid \\
  & Size In & Size of objects in the inner grid \\
  & Color In & Color of objects in the inner grid \\ \hline
  \multirow{4}{*}{2$\times$2Grid} & Num/Pos & Number and position of objects in the grid \\
  & Type & Type of objects in the grid \\
  & Size & Size of objects in the grid \\
  & Color & Color of objects in the grid \\ \hline
  \multirow{4}{*}{3$\times$3Grid} & Num/Pos & Number and position of objects in the grid \\
  & Type & Type of objects in the grid \\
  & Size & Size of objects in the grid \\
  & Color & Color of objects in the grid \\ \hline
  \end{tabular}
\end{table}

Figure \ref{fig:datasets-instance} is an overview of the seven image configurations in RAVEN \citeappendix{zhang2019raven_appendix} and I-RAVEN \citeappendix{hu2021stratified_appendix}. \textit{Center} is the basic configuration with only a single centric object in the image. \textit{L-R} and \textit{U-D} have two components organized with left-right and up-down layouts. The components of \textit{O-IC} and \textit{O-IG} are arranged in terms of the in-out layout. The sample of \textit{2$\times$2Grid} and \textit{3$\times$3Grid} includes a 2$\times$2 or 3$\times$3 object grid. In the datasets, O-IG, 2$\times$2Grid, and 3$\times$3Grid introduce rules on object grids while other configurations only consider rules on a single component. Table \ref{tab:datasets-attribute} introduces the attributes that change with the specific abstract rules in each image configuration. The attribute-changing rules are categorized as follows.
\begin{enumerate}
  \item \textbf{Constant}: the attribute keeps unchanged in rows;
  \item \textbf{Progress}: the attribute increases or decreases with the same stride in rows;
  \item \textbf{Arithmetic}: the attribute of the third image is computed from the attributes of the first two images via specific arithmetic operations (e.g., addition and subtraction operations);
  \item \textbf{Distribution Three}: the attributes in rows are three fixed values in different orders.
\end{enumerate}
Except for the attributes in Table 1, both datasets introduce noisy attributes, which can be randomly sampled in the feasible set of values. For example, the rotation of objects can be stochastically selected in the configurations of non-grid layout and the rotation, color, and position of objects in grids are noisy attributes in the configurations with grid layouts. In our experiments, we remove the noise of the object rotation and color from O-IG, 2$\times$2Grid, and 3$\times$3Grid to reduce the number of possible correct answers.

\section{Experiment Configurations}

\subsection{CRAB}

This section introduces the architectures and hyperparameters of CRAB. In the following part, we will introduce the architectures of $f_{\text{enc}}$, $f_{\text{pair}}^{m}$, $f_{\text{relation}}^{m}$, $f_{\text{target}}^{m}$, and $f_{\text{dec}}$.

\begin{itemize}[leftmargin=0.5cm]
  \item $f_{\text{enc}}$. The encoder $f_{\text{enc}}$ is a convolutional neural network that extracts the mean of concepts. The architecture is
  \begin{itemize} 
      \item 4 $\times$ 4 Conv, stride 2, padding 1, 32 BatchNorm, ReLU
      \item 4 $\times$ 4 Conv, stride 2, padding 1, 64 BatchNorm, ReLU
      \item 4 $\times$ 4 Conv, stride 2, padding 1, 128 BatchNorm, ReLU
      \item 4 $\times$ 4 Conv, stride 2, padding 1, 256 BatchNorm, ReLU
      \item 4 $\times$ 4 Conv, 512 BatchNorm, ReLU
      \item Fully Connected, $M \times d_{z}$
  \end{itemize}
  The output of the encoder is split into the mean of $M$ concepts, and the size of each concept representation is $d_{z}$.
  \item $f_{\text{pair}}^{m}$. The input of $f_{\text{pair}}^{m}$ is a pair of concept representations. The architecture is
  \begin{itemize} 
      \item Fully Connected, 512 ReLU
      \item Fully Connected, 512 ReLU
      \item Fully Connected, 64
  \end{itemize}
  \item $f_{\text{relation}}^{m}$. We concatenate all the pair representations as the input of $f_{\text{relation}}^{m}$, and the architecture of the network is
  \begin{itemize} 
      \item Fully Connected, 2048 ReLU
      \item Fully Connected, 1024 ReLU
      \item Fully Connected, 512 ReLU
      \item Fully Connected, $2d_{r}$
  \end{itemize}
  where $d_{r}$ is the size of rule latent variables. The output of size $2d_{r}$ is split into the mean and standard deviation of $\boldsymbol{r}^{m}$.
  \item $f_{\text{target}}^{m}$. We adopt a fully convolutional network to predict the means of target latent concepts from $\boldsymbol{Z}^{m}$:
  \begin{itemize} 
    \item 3 $\times$ 3 Conv, stride 1, padding 1, 128 ReLU
    \item 3 $\times$ 3 Conv, stride 1, padding 1, 128 ReLU
    \item 3 $\times$ 3 Conv, stride 1, padding 1, 128 ReLU
    \item 3 $\times$ 3 Conv, stride 1, padding 1, $d_{z}$ ReLU
  \end{itemize}
  We set the kernel size as 3$\times$3, stride as 1, and padding size as 1 for the convolutional layers to keep the shape of $\boldsymbol{Z}^{m}$.
  \item $f_{\text{dec}}$. The decoder accepts the representations of $M$ concepts as input and outputs the mean of pixel values for target images. The architecture is
  \begin{itemize} 
      \item 1 $\times$ 1 Deconv, 128 BatchNorm, LeakyReLU
      \item 4 $\times$ 4 Deconv, 64 BatchNorm, LeakyReLU
      \item 4 $\times$ 4 Deconv, stride 2, padding 1, 64 BatchNorm, LeakyReLU
      \item 4 $\times$ 4 Deconv, stride 2, padding 1, 32 BatchNorm, LeakyReLU
      \item 4 $\times$ 4 Deconv, stride 2, padding 1, 32 BatchNorm, LeakyReLU
      \item 4 $\times$ 4 Deconv, stride 2, padding 1, 1 Sigmoid
  \end{itemize}
  The negative slope of LeakyReLU is $0.02$, and we use the Sigmoid activation function to scale the output pixel values into $(0,1)$.
\end{itemize}
\begin{table}[t]
    \caption{Configuration-specific hyperparameters of CRAB.}
    \label{tab:hyperparameter-crab}
    \centering
    \begin{tabular}{c|ccccccc}
    \hline
    Hyperparameter & Center & L-R & U-D & O-IC & O-IG & 2$\times$2Grid & 3$\times$3Grid \\
    \hline
    $\beta_{r}$ & 10 & 5 & 5.5 & 6 & 3 & 3 & 8 \\
    $\beta_{t}$ & 10 & 5 & 5.5 & 3 & 3 & 3 & 8 \\
    $\sigma_{z}$ & 0.3 & 0.1 & 0.1 & 0.4 & 0.1 & 0.3 & 0.3 \\
    $M$ & 4 & 8 & 8 & 6 & 8 & 8 & 10 \\
    \hline
    \end{tabular}
\end{table}
We set $K=4$, learning rate as $3 \times 10^{-4}$, batch size as 512, $d_{z}=32$, $d_{r}=2$, and $\beta_{\text{cls}} = 1 \times 10^{-5}$. The parameters are updated with the RMSprop \citeappendix{hinton2012neural_appendix} optimizer. See Table \ref{tab:hyperparameter-crab} for other configuration-specific hyperparameters.

\subsection{Selective Solvers}

We compare CRAB with representative selective solvers \citeappendix{hu2021stratified_appendix, barrett2018measuring_appendix, zheng2019abstract_appendix, zhang2019learning_appendix, wu2020scattering_appendix} to illustrate the performance difference between the generative and selective solvers in the RAVEN and I-RAVEN datasets. We follow the existing implementations in code repositories and use the recommended hyperparameters of the selective solvers in the experiments.

\subsection{Bottom-Right Generative Solvers}

We take the official implementation and recommended configurations of the bottom-right generative solvers in the experiments \citeappendix{zhang2021learning_appendix, zhang2021abstract_appendix, pekar2020generating_appendix}. We find that ALANS \citeappendix{zhang2021learning_appendix} is unstable if trained with randomly initialized parameters. Therefore, we initialize the parameters of ALANS with the checkpoint provided by the authors. Since the auxiliary loss in the official code of GCA \citeappendix{pekar2020generating_appendix} can only handle the PGM dataset \citeappendix{barrett2018measuring_appendix}, we modify the output size of the auxiliary network to adapt to the rule annotations in RAVEN and I-RAVEN.

\subsection{Aribitrary-Position Generative Solvers}

\begin{table}[t]
  \caption{Configuration-specific hyperparameters of the baselines.}
  \label{tab:hyperparameter-compared-model}
  \centering
  \begin{tabular}{c|c|ccccccc}
  \hline
  Model & Hyperparameter & Center & L-R & U-D & O-IC & O-IG & 2$\times$2Grid & 3$\times$3Grid \\
  \hline
  \multirow{1}{*}{ANP} & learning rate & $5 \times 10^{-5}$ & $1 \times 10^{-5}$ & $1 \times 10^{-5}$ & $5 \times 10^{-6}$ & $5 \times 10^{-6}$ & $3 \times 10^{-5}$ & $3 \times 10^{-5}$ \\
  \hline
  \multirow{1}{*}{LGPP} & number of concepts & 5 & 10 & 10 & 10 & 10 & 10 & 10 \\
  \hline
  \multirow{5}{*}{CLAP-NP} & number of concepts & 5 & 10 & 10 & 6 & 8 & 8 & 10 \\
  & $\beta_{t}$ & 100 & 50 & 50 & 30 & 30 & 30 & 80 \\
  & $\beta_{f}$ & 100 & 50 & 50 & 60 & 30 & 30 & 80 \\
  & $\beta_{TC}$ & 100 & 50 & 50 & 50 & 30 & 30 & 80 \\
  & $\sigma_{z}$ & 0.1 & 0.1 & 0.1 & 0.4 & 0.1 & 0.3 & 0.3 \\ \hline
  \end{tabular}
\end{table}

We employ aribitrary-position generative solvers Transformer \citeappendix{vaswani2017attention_appendix}, ANP \citeappendix{kim2019attentive_appendix}, LGPP \citeappendix{shi2021raven_appendix}, and CLAP-NP \citeappendix{shi2022compositional_appendix} as the baseline models. Transformer conducts deterministic conditional generation to predict answers at arbitrary positions. ANP explicitly parses rule latent variables on RPMs and predicts target images through stochastic conditional generation. LGPP and CLAP-NP hold the same idea to decompose an image into interpretable concepts but can hardly solve arbitrary-position answer generation on RPMs with discrete attributes and rules. We utilize the encoder and decoder of CRAB to extract representations and Transformer for target prediction in the low-dimensional space. We set the learning rate as $1 \times 10^{-4}$, the size of representations as $256$, the number of Transformer blocks as $4$, the number of attention heads as $4$, the hidden size of feedforward networks as $1024$, and the dropout as $0.1$. The parameters of Transformer are updated by the Adam \citeappendix{kingma2014adam_appendix} optimizer. For ANP, we set the size of global latent variables as $1024$ and the batch size as $512$. The configuration-specific hyperparameters are given in Table \ref{tab:hyperparameter-compared-model}, and the remaining hyperparameters refer to the 2D regression configuration in \citeappendix{kim2019attentive_appendix}. For LGPP, we set the learning rate as $5 \times 10^{-4}$ and the batch size as 256, the size of axis latent variables as 4, the size of axis representations as 4, and the input size of the RBF kernel as 8. To convert axis latent variables to axis representations, we adopt an MLP with hidden sizes [64, 64]. The MLP that extracts the input features of RBF kernels has hidden sizes [128, 128, 128, 128]. The weight $\beta$ in the ELBO is 10. Other configuration-specific hyperparameters are given in Table \ref{tab:hyperparameter-compared-model}. For CLAP-NP, we follow the CRPM configuration in the official repository while adjusting the learning rate to $5 \times 10^{-4}$, the batch size to 256, and the concept size to 8. See Table \ref{tab:hyperparameter-compared-model} for other configuration-specific hyperparameters.

\subsection{Computational Resource}

We conduct the training and evaluation of the models on the server with Intel(R) Xeon(R) Platinum 8375C CPUs, 24GB NVIDIA GeForce RTX 3090 GPUs, 512GB RAM, and Ubuntu 18.04.6 LTS. CRAB is implemented with PyTorch \citeappendix{paszke2019pytorch_appendix}.

\section{Additional Experimental Results}

In this section, we will provide additional experimental results on different configurations. In the experiment of arbitrary-position answer generation, we display the results of predicting multiple target images ($|T|=2$). And in the other experiments, we provide the results of the image configurations not given in the main text.

\subsection{Arbitrary-Position Answer Generation}

\begin{figure}[t]
  \centering
  \includegraphics[width=\textwidth]{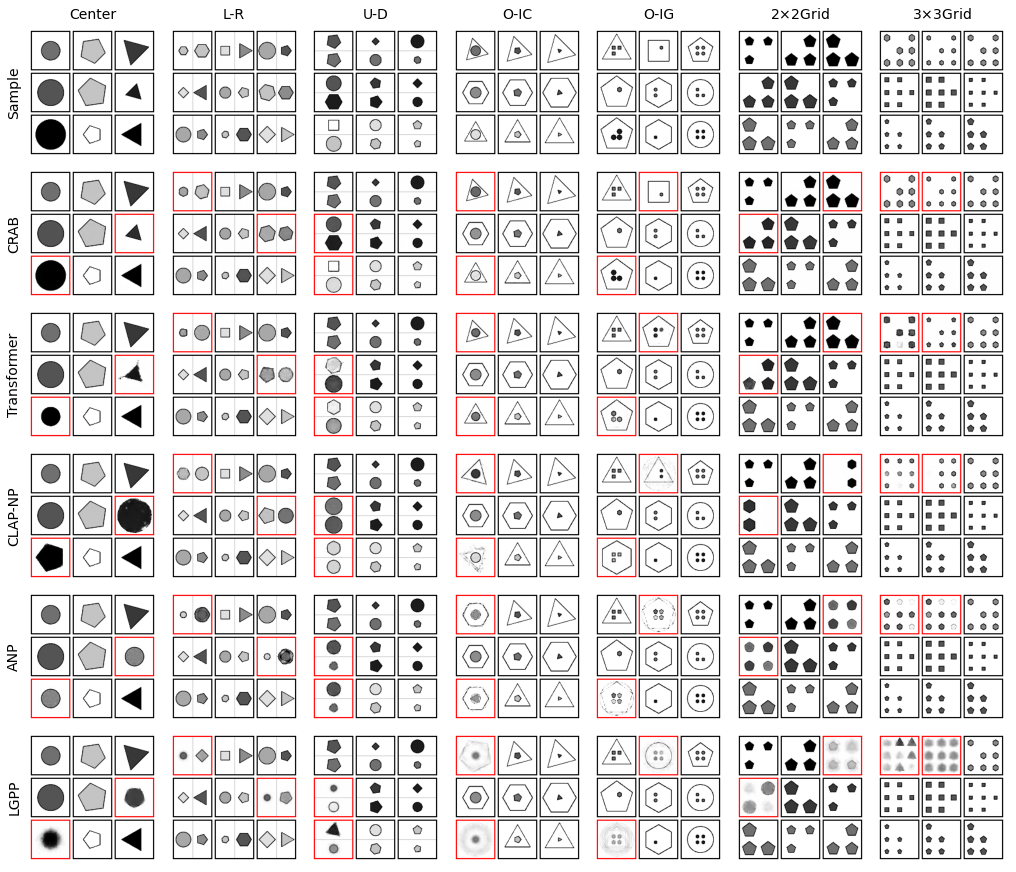}
  \caption{\textbf{Arbitrary-position answer generation on RAVEN ($|T|=2$)}. The original RPM samples are displayed in the first row, and the prediction results are highlighted with red boxes in each RPM.}
  \label{fig:exp-arb-pos-ans-gen}
\end{figure}

In this experiment, Figure \ref{fig:exp-arb-pos-ans-gen} shows the prediction results of CRAB and the compared models to illustrate our model's ability in multiple-position answer generation. The difficulty of this task lies in that the models are trained to predict only one target image in RPMs and tested to generate multiple answers in a matrix. We argue that the generalization of the abstract reasoning process is the key to applying the models in novel situations. Compared with the baselines, CRAB generates the most accurate answers in all the configurations while the generation results of Transformer usually deviate from the original samples, e.g., the generated target images of O-IC have incorrect colors. Additionally, CLAP-NP and ANP produce clear target images, but they can hardly meet the underlying rules in the problem panel. LGPP generates ambiguous answers in RPMs with non-continuous attributes and rules. Visualizing multiple-position answer generation can help us verify the generalization ability of models when solving RPMs in novel problem configurations.

\subsection{Concept Learning}

\begin{figure}[t]
  \centering
  \includegraphics[width=\textwidth]{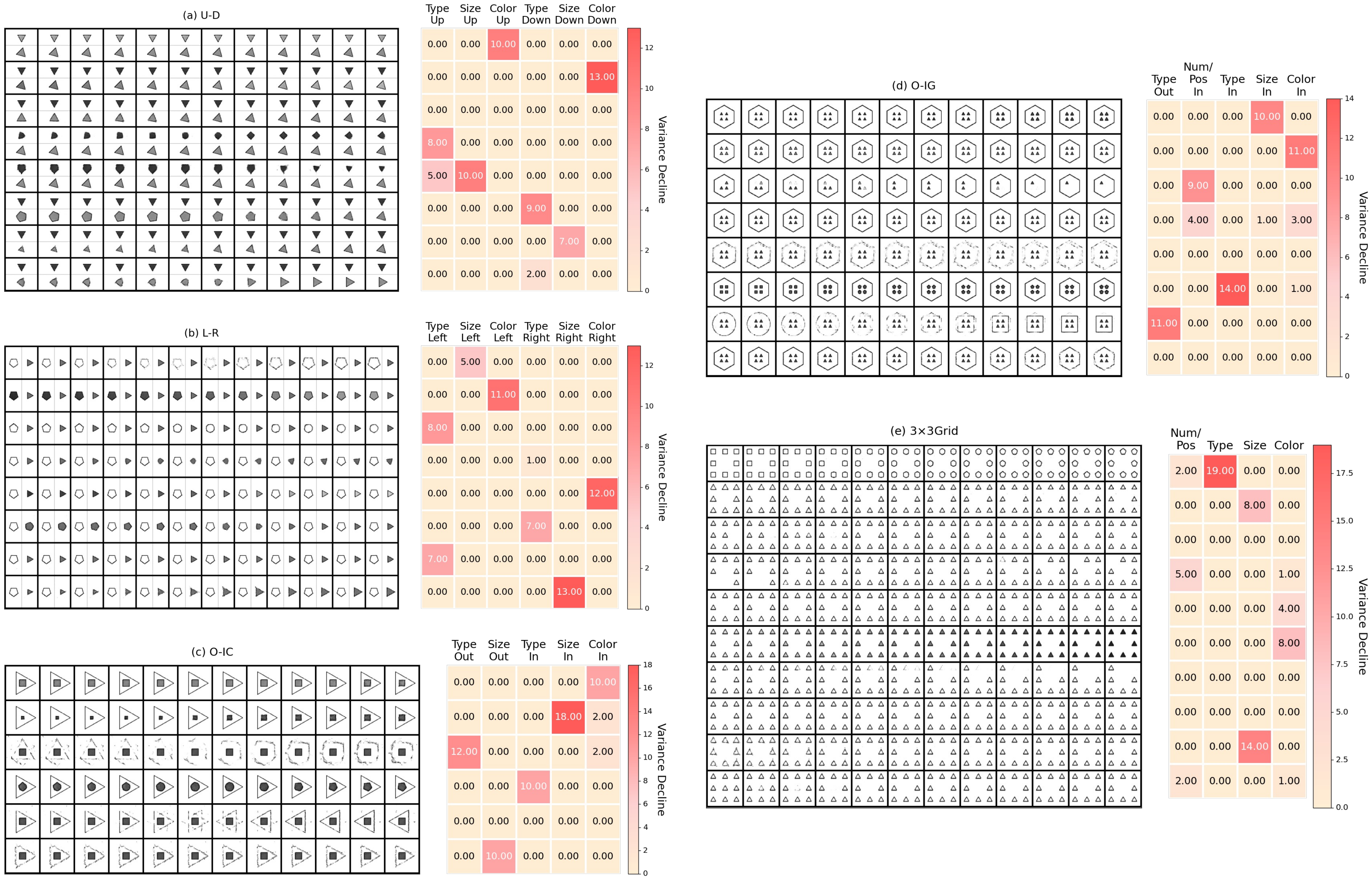}
  \caption{\textbf{Additional experimental results of concept learning}. Figures (a)-(e) display the results of U-D, L-R, O-IC, O-IG, and 3$\times$3Grid. In each plot, the left panel is the interpolation results on different latent concepts and the right panel displays variance decline scores between the learned concepts and real changeable attributes.}
  \label{fig:exp-concept-learning}
\end{figure}

As mentioned in the main text, concept learning is the foundation of interpretability because CRAB parses and abstracts in terms of concept-specific rules. Therefore, the quality of concept learning will influence the results of global rule abstraction and answer selection. This experiment illustrates the meaning of each concept by interpolating concept representations and visualizing variance declines (VDs). Here we provide the results of U-D, L-R, O-IC, O-IG, and 3$\times$3Grid in Figure \ref{fig:exp-concept-learning} as the addition to the main text. We find that the concept learned from the datasets can interpret different attributes of images in most cases, and it is straightforward to make out the correspondence between concepts and attributes through the VDs. The attribute \textit{Num/Pos} in the configurations with object grids is usually encoded by more than one concept, e.g., \textit{Num/Pos In} in O-IG is represented in the third and fourth concepts according to the VDs. The additional experimental results of concept learning will further account for the interpretability of CRAB in abstract reasoning.

\subsection{Interpretable Answer Selection} 

\begin{figure}[t]
  \centering
  \includegraphics[width=0.9\textwidth]{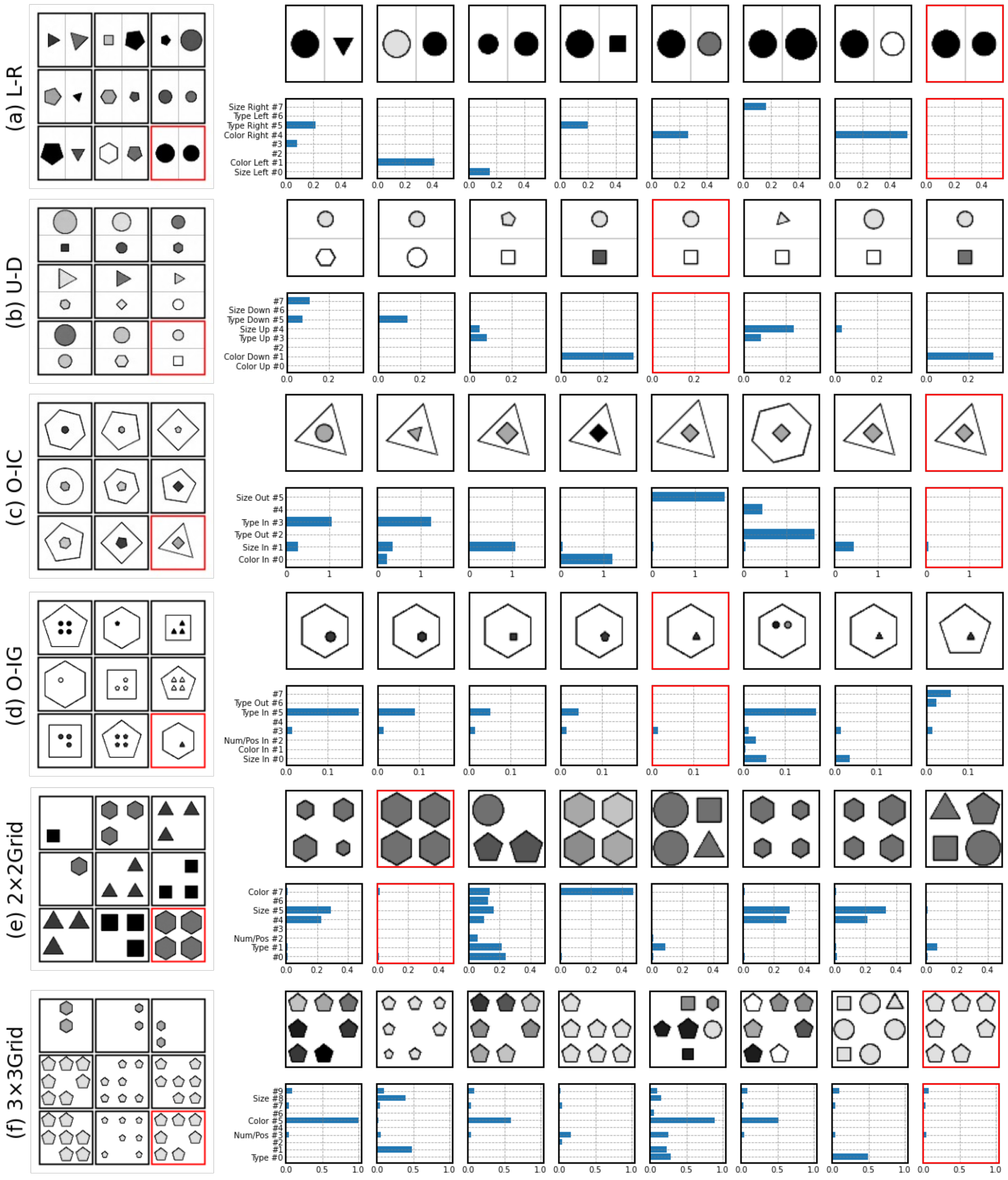}
  \caption{\textbf{The answer selection process of CRAB}. Figures (a)-(f) visualize the candidate images and the concept-level distance between the prediction and candidates. We annotate the corresponding attribute of each concept to illustrate the interpretability of answer selection.}
  \label{fig:exp-interpretable-prediction-appendix}
\end{figure}

Based on concept learning, CRAB can interpret the selection and exclusion of candidates through concept-wise distances between the prediction result and candidates. Figure \ref{fig:exp-interpretable-prediction-appendix} provides concept-wise distances on samples from L-R, U-D, O-IC, O-IG, 2$\times$2Grid, and 3$\times$3Grid as additional results to illustrate the interpretability of answer selection. For non-grid configurations L-R, U-D, and O-IC, the distances can exactly reflect the incorrect attributes in distractors. In the O-IG, 2$\times$2Grid, and 3$\times$3Grid configurations, the distances are sometimes confounded if the distractors have multiple incorrect attributes (e.g., the third, fifth, and sixth candidates in the instance of 3$\times$3Grid). But overall, the concept-wise distances still reflect the correctness of candidates.

\subsection{Global Rule Abstraction}

\begin{figure}[t]
  \centering
  \includegraphics[width=\textwidth]{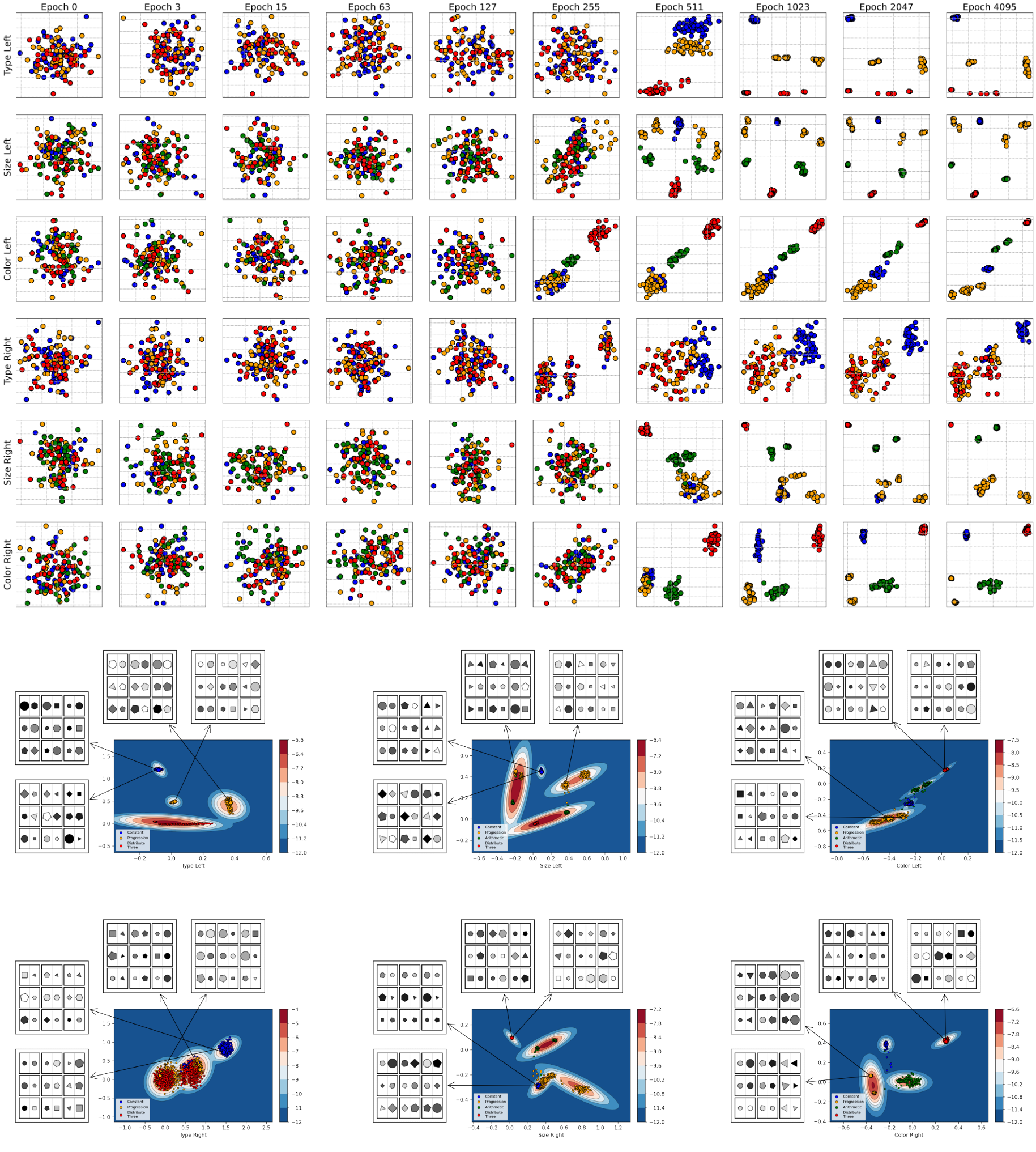}
  \caption{\textbf{Visualization results of global rule abstraction on L-R}. The top panel displays the change of rule distributions on concepts where the scatters represent the rule latent variables and the colors of scatters indicate the actual categories of rules. The bottom panel visualizes the global rules where the contours in the background show the logarithmic probability density $\log p(\boldsymbol{r}^{m})$. We randomly select scatters from the clusters and visualize the corresponding RPM panels to illustrate the global rules we learned.}
  \label{fig:exp-global-rule-lr}
\end{figure}

\begin{figure}[t]
  \centering
  \includegraphics[width=\textwidth]{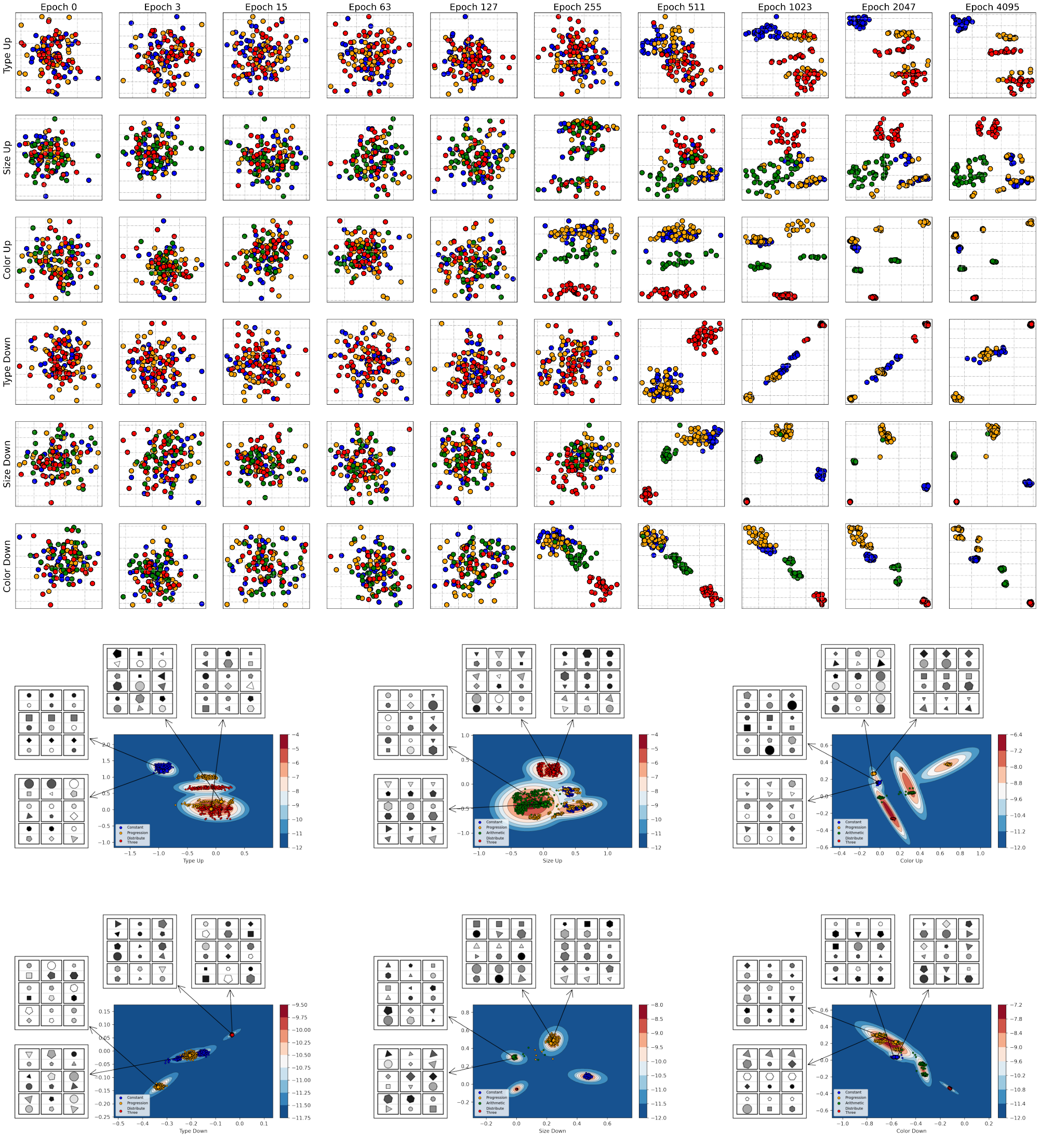}
  \caption{Visualization results of global rule abstraction on U-D.}
  \label{fig:exp-global-rule-ud}
\end{figure}

\begin{figure}[t]
  \centering
  \includegraphics[width=\textwidth]{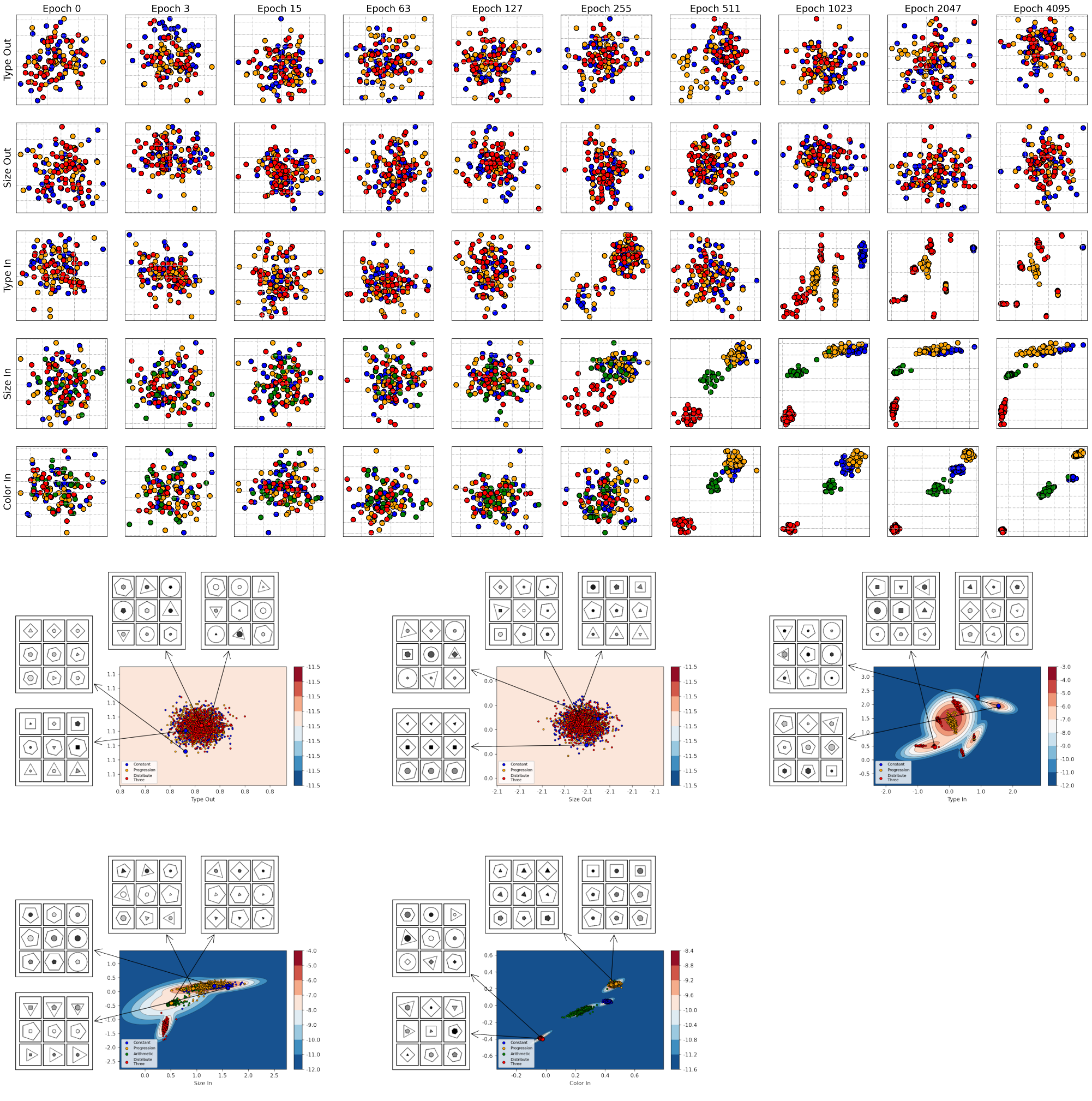}
  \caption{Visualization results of global rule abstraction on O-IC.}
  \label{fig:exp-global-rule-oic}
\end{figure}

\begin{figure}[t]
  \centering
  \includegraphics[width=\textwidth]{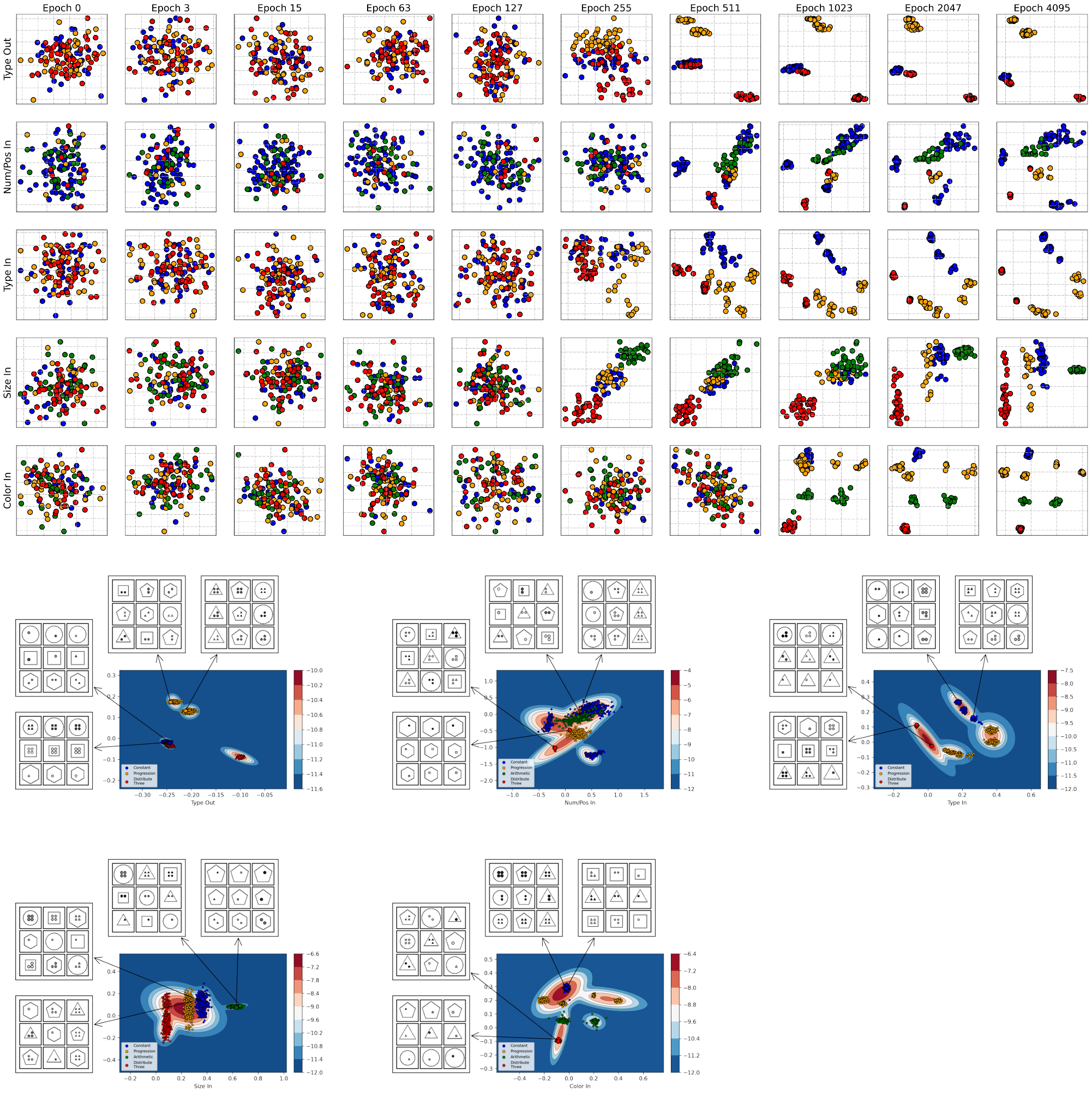}
  \caption{The visualization of global rule abstraction on O-IG.}
  \label{fig:exp-global-rule-oig}
\end{figure}

\begin{figure}[t]
  \centering
  \includegraphics[width=\textwidth]{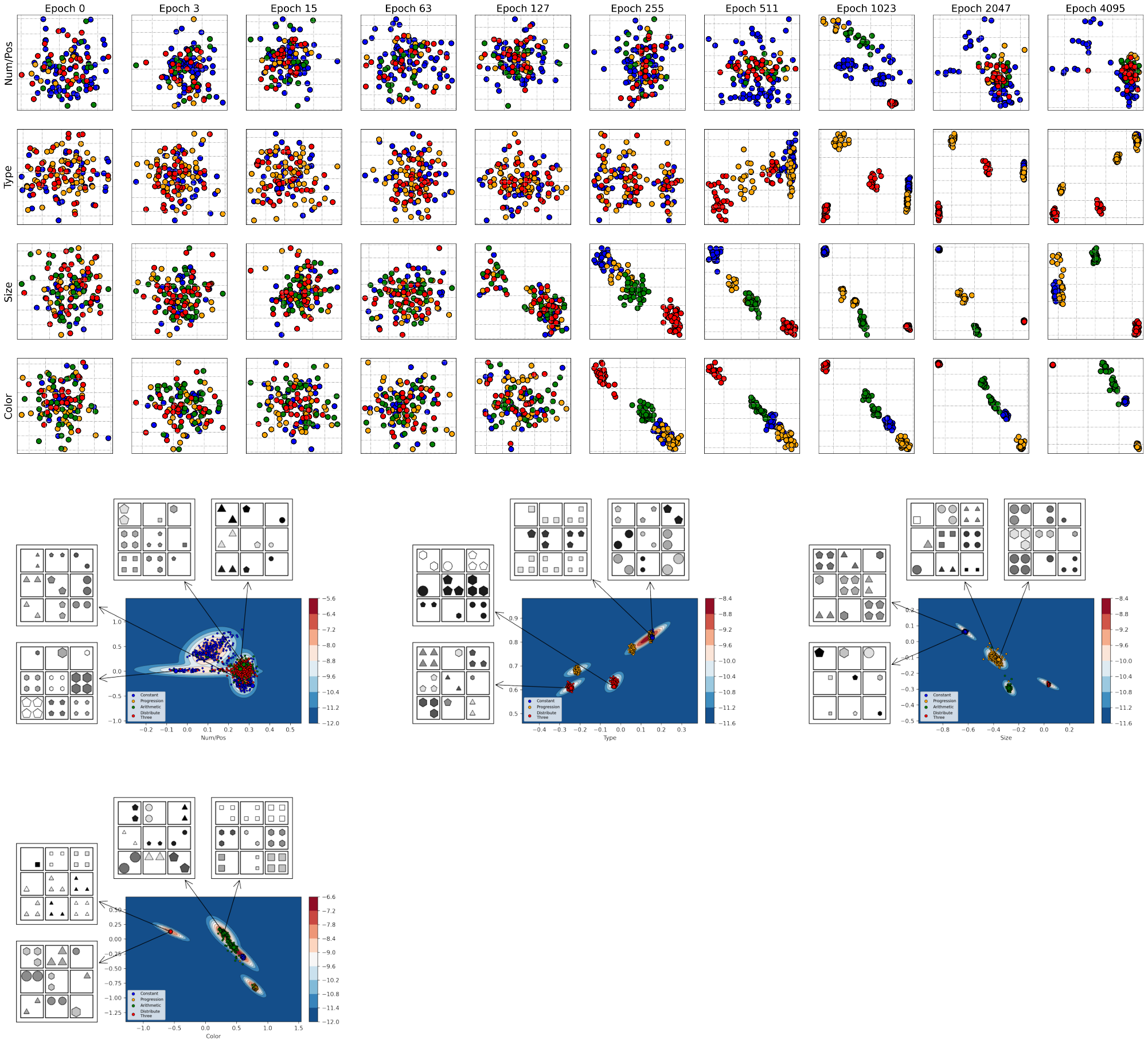}
  \caption{Visualization results of global rule abstraction on 2$\times$2Grid.}
  \label{fig:exp-global-rule-22}
\end{figure}

\begin{figure}[t]
  \centering
  \includegraphics[width=\textwidth]{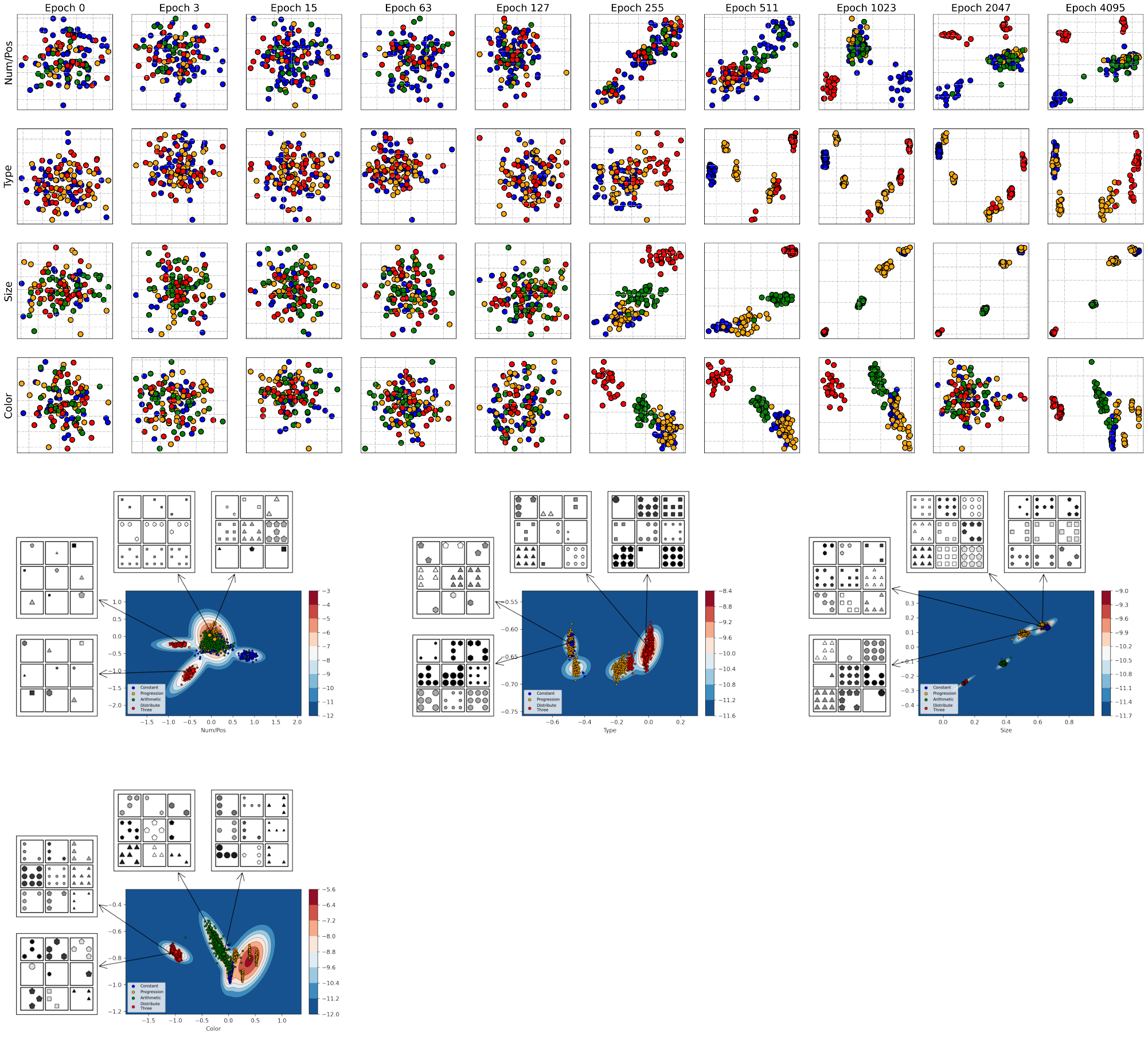}
  \caption{Visualization results of global rule abstraction on 3$\times$3Grid.}
  \label{fig:exp-global-rule-33}
\end{figure}

CRAB achieves the interpretability of rule parsing by automatically discovering the global concept-specific rules shared on the dataset. Figures \ref{fig:exp-global-rule-lr}, \ref{fig:exp-global-rule-ud}, \ref{fig:exp-global-rule-oic}, \ref{fig:exp-global-rule-oig}, \ref{fig:exp-global-rule-22}, and \ref{fig:exp-global-rule-33} illustrate the rule abstraction ability of CRAB in L-R, U-D, O-IC, O-IG, 2$\times$2Grid, 3$\times$3Grid respectively. As shown at the top of the figures, the rule latent variables of the RPMs with the same rule will be close to each other. Sometimes, CRAB splits the RPMs of the same category into different clusters, e.g., in the samples of L-R and O-IG, RPMs having the rule \textit{Progression} on \textit{Type Right} and \textit{Type Out} are divided into two clusters. The visualized RPM panels from the two clusters explain why CRAB further decomposes the rule \textit{Progression}. The first cluster contains the samples where the three rows have the same progressively changed attributes (e.g., [[1, 2, 3], [1, 2, 3], [1, 2, 3]]), and the second cluster contains those have different progressive sequences (e.g., [[1, 2, 3], [2, 3, 4], [3, 4, 5]]). Since CRAB is trained without any annotations of rules, we argue that it is reasonable to categorize the global abstract rules differently. The key point is that CRAB can build and update prior knowledge of rules without auxiliary supervision, and such ability emerges in all image configurations of the RAVEN dataset. We also observe some failure cases in abstracting global rules, i.e., in the attributes \textit{Type Out} and \textit{Size Out} of O-IC, all the samples are allocated into a single cluster, making it difficult to develop priors of concept-specific rules.

\ifCLASSOPTIONcaptionsoff
  \newpage
\fi

\bibliographyappendix{appendix}
\bibliographystyleappendix{IEEEtran}

\end{document}